\documentclass[preprint,review,12pt]{elsarticle}

\usepackage{upgreek}

\newcommand{\CSND}{\texttt{NeurADP+DDQN}}
\newcommand{\CSNF}{\texttt{NeurADP+Fixed}}
\newcommand{\CSGD}{\texttt{Greedy+DDQN}}
\newcommand{\CSGF}{\texttt{Greedy+Fixed}}

\newcommand{\CSCrowdShipperSet}{\mathcal{C}}
\newcommand{\CSOrderRequestSet}{\mathcal{O}}
\newcommand{\CSPricingDecisionSet}{\mathcal{P}}
\newcommand{\CSPossibleBatchesSet}{\mathcal{B}}
\newcommand{\CSPossibleRoutesSet}{\mathcal{R}}
\newcommand{\CSFeasibleMatchingSet}{\mathcal{F}}

\newcommand{\CSAllActionsSet}{\mathbf{A}}
\newcommand{\CSPermutationSet}{\mathcal{S}}
\newcommand{\CSEpisodeSet}{\texttt{$N$}}

\newcommand{\CSCrowdShipperState}{{\texttt{$C$}}}
\newcommand{\CSRequestState}{{\texttt{$O$}}}
\newcommand{\CSIndividualCrowdShipper}{{\texttt{$c$}}}
\newcommand{\CSCrowdShipperHome}{{\text{home}}}
\newcommand{\CSIndividualOrderRequest}{{\texttt{$o$}}}
\newcommand{\CSRequestDestination}{{\text{destination}}}
\newcommand{\CSRequestEpochsRemaining}{{\text{epochs}}}
\newcommand{\CSRequestDeadline}{{\text{deadline}}}
\newcommand{\CSDeadlineTime}{{\text{D}}}
\newcommand{\CSEpochLength}{\texttt{$\delta$}}
\newcommand{\CSCurrentTime}{\texttt{$t$}}
\newcommand{\CSStateNote}{\texttt{$S$}}
\newcommand{\CSTime}{\mathcal{T}}

\newcommand{\CSLocations}{\texttt{$\mathcal{L}$}}
\newcommand{\CSTimeIntervals}{\texttt{$\delta$}}

\newcommand{\CSTravelTime}{\texttt{time}}
\newcommand{\CSMaxLocations}{\kappa}
\newcommand{\CSSingleBatch}{\texttt{$b$}}
\newcommand{\CSRoute}{\texttt{$r$}}
\newcommand{\CSOrderPermutation}{\texttt{$\sigma$}}
\newcommand{\CSTotalTravelTime}{\texttt{RouteTime}}
\newcommand{\CSBestRoute}{\CSRoute^\star}
\newcommand{\CSBaseCompensation}{\texttt{BaseFee}}
\newcommand{\CSDetourCompensation}{\texttt{DetourFee}}
\newcommand{\CSDetourTime}{\texttt{DetourTime}}
\newcommand{\CSMatchCost}{\texttt{MatchCost}}
\newcommand{\CSDelayCost}{\texttt{DelayCost}}
\newcommand{\CSLostCost}{\texttt{LostCost}}
\newcommand{\CSExpectedCost}{\texttt{ExpMatchCost}}

\newcommand{\CSAdjustedBaseFee}{\texttt{AdjBaseFee}}

\newcommand{\CSAcceptProbFunc}{\psi}

\newcommand{\CSCostName}{\texttt{R}}

\newcommand{\CSPostDecisionState}{\CSStateNote^\texttt{Post}}
\newcommand{\CSOrderPost}{\CSRequestState^\texttt{Post}}
\newcommand{\CSOrderRetained}{\CSRequestState^\texttt{Retained}}

\newcommand{\CSpost}{\texttt{statepost}}
\newcommand{\CSnext}{\texttt{statenext}}

\newcommand{\CSCrowd}{\texttt{Crowd}}
\newcommand{\CSOrd}{\texttt{Order}}
\newcommand{\CSAcc}{\texttt{Accept}}

\newcommand{\CSValueNote}{V}
\newcommand{\CSValueName}{\hat{\texttt{v}}}

\newcommand{\CSEpisode}{\texttt{$n$}}

\newcommand{\CSPricingIndividualMultiplier}{p}
\newcommand{\CSActionVar}{x}
\newcommand{\CSDelayVar}{y}
\newcommand{\CSActionVector}{\textbf{\CSActionVar}_\CSCurrentTime}
\newcommand{\CSDelayVector}{\textbf{\CSDelayVar}_\CSCurrentTime}
\newcommand{\CSPricingVector}{\textbf{\CSPricingIndividualMultiplier}_\CSCurrentTime}
\newcommand{\CSDecisionsVector}{\textbf{a}_\CSCurrentTime}

\newcommand{\CSExogenousInformation}{W}

\newcommand{\CSPricingDistribution}{P}
\newcommand{\CSSmallPolicy}{\pi}
\newcommand{\CSBigPolicy}{\Pi}

\usepackage[utf8]{inputenc}
\usepackage{fullpage,times}
\usepackage{amsmath,amssymb}
\usepackage{mathtools}
\usepackage{graphicx}
\usepackage{algorithm}
\usepackage{algorithmicx}
\usepackage{algpseudocode}
\usepackage{subfig}
\usepackage{caption}
\usepackage{url}
\usepackage[numbers]{natbib} % bibliography,references (options: round, numbers)
\usepackage{placeins}
\usepackage{siunitx}
\usepackage{url}
\usepackage{xspace}
\usepackage[table,xcdraw]{xcolor}

\usepackage{hyperref}
\hypersetup{
colorlinks={true},
linkcolor={Blue3},
urlcolor={Blue3},
citecolor={Blue3},
pdfstartview={FitH},
pdfauthor={Author},
pdftitle={Title},
pdfsubject={Subject}
plainpages=false
}

\usepackage{enumitem}

\usepackage{booktabs} % table toprule etc
\usepackage{multirow}
\usepackage{multicol}
\usepackage{makecell}
\usepackage{amsmath}

\usepackage[toc,page]{appendix}

\usepackage{setspace}
\makeatletter
\def\ps@pprintTitle{%
  \let\@oddhead\@empty
  \let\@evenhead\@empty
  \def\@oddfoot{\reset@font\hfil\thepage\hfil}
  \let\@evenfoot\@oddfoot
}
\makeatother

\definecolor{wisconsin-red}{rgb}{0.6,0,0}
\definecolor{darkgreen}{rgb}{0.2,0.6,0.2}
\definecolor{maroon}{rgb}{0.5, 0.0, 0.0}
\definecolor{violet}{rgb}{0.75, 0.0, 1.0}
\definecolor{lightgray}{gray}{0.9}
\definecolor{navyblue}{rgb}{0.0, 0.0, 0.5}
\definecolor{darkmidnightblue}{rgb}{0.0, 0.2, 0.4}
\definecolor{midnightblue}{rgb}{0.0,0.4,0.85}
\definecolor{Gray}{gray}{0.75}
\definecolor{darkgreen}{rgb}{0,0.5,0}
\definecolor{apricot}{rgb}{0.98, 0.81, 0.69}

\newcolumntype{C}[1]{>{\centering\arraybackslash}p{#1}}
\newcolumntype{P}[1]{>{\raggedright\arraybackslash}p{#1}}
\newcolumntype{L}[1]{>{\raggedleft\arraybackslash}p{#1}}

\journal{}

\begin{document}\sloppy
\setlength{\parindent}{2em}

\begin{frontmatter}
\title{Joint Matching and Pricing for Crowd-shipping with In-store Customers} % tentative title

\author[1]{Arash Dehghan}
\ead{arash.dehghan@torontomu.ca}

\author[1]{Mucahit Cevik\corref{cor1}%
\fnref{fn1}}
\ead{mcevik@torontomu.ca}

\author[2]{Merve Bodur}
\ead{merve.bodur@ed.ac.uk}

\author[3]{Bissan Ghaddar}
\ead{bghaddar@ivey.ca}

\cortext[cor1]{Corresponding author}
\fntext[fn1]{Toronto Metropolitan University, Toronto, ON, Canada}
\address[1]{Toronto Metropolitan University, Toronto, ON, Canada}
\address[2]{School of Mathematics and Maxwell Institute for Mathematical Sciences, University of Edinburgh, Edinburgh, UK}
\address[3]{Western University, London, ON, Canada}

%%%%%%%%%%%%%%%%%%%%%%%%%%%%%%%%%%%%%%%%%%%%%%%%%%%%%%%%%%%%%%%%%%%%%%%%
%%%%%%%%%%%%%%%%%%%%%%%%%%%%%%%%%%%%%%%%%%%%%%
%%%%%%%%%%%%% Section: Abstract %%%%%%%%%%%%%%
%%%%%%%%%%%%%%%%%%%%%%%%%%%%%%%%%%%%%%%%%%%%%%
\begin{abstract}
This paper examines the use of in-store customers as delivery couriers in a centralized crowd-shipping system, targeting the growing need for efficient last-mile delivery in urban areas. We consider a brick-and-mortar retail setting where shoppers are offered compensation to deliver time-sensitive online orders. To manage this process, we propose a Markov Decision Process (MDP) model that captures key uncertainties, including the stochastic arrival of orders and crowd-shippers, and the probabilistic acceptance of delivery offers. Our solution approach integrates Neural Approximate Dynamic Programming (NeurADP) for adaptive order-to-shopper assignment with a Deep Double Q-Network (DDQN) for dynamic pricing. This joint optimization strategy enables multi-drop routing and accounts for offer acceptance uncertainty, aligning more closely with real-world operations. Experimental results demonstrate that the integrated NeurADP + DDQN policy achieves notable improvements in delivery cost efficiency, with up to 6.7\% savings over NeurADP with fixed pricing and approximately 18\% over myopic baselines. We also show that allowing flexible delivery delays and enabling multi-destination routing further reduces operational costs by 8\% and 17\%, respectively. These findings underscore the advantages of dynamic, forward-looking policies in crowd-shipping systems and offer practical guidance for urban logistics operators.
\end{abstract}
\begin{keyword}
Crowd-shipping\sep ADP\sep NeurADP \sep Value function approximation
\end{keyword}
\end{frontmatter}
\vspace{-8pt}
%%%%%%%%%%%%%%%%%%%%%%%%%%%%%%%%%%%%%%%%%%%%%%%%%%%%%%%%
%%%%%%%%%%%%% Section: Introduction %%%%%%%%%%%%%%
%%%%%%%%%%%%%%%%%%%%%%%%%%%%%%%%%%%%%%%%%%%%%%%%%%%%%%%%
\section{Introduction}

The explosive growth of e-commerce, accelerated by changing consumer habits, technological innovation, and the COVID-19 pandemic, has reshaped retail and sharply increased online order volumes. Global e-commerce sales are expected to surpass \$8 trillion by 2027, which represents a 39\% rise from 2023, and this increase is driving a surge in last-mile deliveries. This growth is projected to cause a 61\% increase in global last-mile delivery vehicles by 2030, exacerbating urban congestion, parking shortages, and commute delays. For example, between 2018 and 2023, congestion rose by 21\% in Dublin and 15\% in Milan, while delivery traffic in Bengaluru is expected to add 7 minutes to a typical 10 km commute by 2030. Environmental consequences are also significant, with delivery emissions forecasted to grow 60\% and contribute 13\% of urban emissions \citep{frobel2024sustainable}. Crowd-shipping offers a promising solution by utilizing non-professional local couriers to fulfill deliveries. This model reduces dependence on fleet vehicles, cutting emissions, easing congestion, and improving logistics efficiency. With delivery costs and delays contributing to 48\% and 23\% of online cart abandonment, respectively~\citep{baymard2024cart}, crowd-shipping lowers per-parcel costs by up to 29\% and minimizes fixed overhead. It also brings environmental and social benefits by integrating deliveries into existing travel patterns, potentially reducing vehicle kilometers by 17\% and mitigating urban pollution \citep{allen2010sustainability, zhang2023exploring}. Driven by these advantages, the global crowdsourced logistics market, valued at \$12.5 billion in 2023, is projected to reach \$37.5 billion by 2032, with a 12.5\% annual growth rate~\citep{dataintelo2024crowdsourced}. Retailers are rapidly adopting this model, with 90\% expected to use crowdsourced delivery for certain orders by 2028 and 40\% planning to offer two-hour delivery windows \citep{roadie2018crowdsourced}. Platforms like Amazon Flex and GoShare demonstrate the model's success. Amazon Flex relies on independent drivers using personal vehicles \citep{amazonflex}, while GoShare connects businesses with local couriers through an app, serving clients such as Costco, Pier 1 Imports, and Ace Hardware \citep{goshare2017vision}. The widespread adoption of crowd-shipping signals its growing role in solving last-mile delivery challenges.

In this paper, we consider a centralized crowd-shipping setting in which a single brick-and-mortar store leverages in-store customers (i.e., individuals already present at the store and planning to travel home) as potential crowd-shippers for fulfilling incoming online orders. This concept is exemplified by a program implemented by Walmart, where in-store shoppers deliver online orders on their way home \citep{dayarian2020crowdshipping}. These online orders are desired to be delivered within a short time window (e.g., a few hours), and any orders that remain unserved past their delivery deadline are either canceled or fulfilled using traditional last-mile delivery methods at a higher cost to the store. At each decision point, the store selects a subset of outstanding orders and offers them to eligible in-store customers, along with a proposed compensation amount for completing the delivery. Each customer may choose to accept or reject the offer based on the payment offered and detour required for them to make such deliveries. Consequently, the system must operate under uncertainty regarding the arrival of both online orders and eligible crowd-shippers, as well as their willingness to accept delivery offers, while aiming to minimize the store's total operational cost. Thus, two critical decision components define the performance of such a system. The first is the matching and timing strategy, which involves determining which orders to assign to which in-store crowd-shippers and when to delay assignment in hopes of a better future match. This component requires balancing the urgency of deliveries against the stochastic nature of future shopper arrivals. The second is the pricing and incentive mechanism which decide how much compensation to offer each in-store customer to increase the likelihood that they accept a delivery task. Since shoppers are not obligated to accept delivery requests, offering too little may lead to rejection and unserved orders, while offering too much erodes the cost benefits of crowd-shipping.

For centralized stores managing crowd-shipping deliveries, adopting a strategic rather than ad hoc approach is essential. Unlike decentralized gig platforms, store-managed systems can optimize assignments by factoring in delivery urgency, shopper routes, order locations, and availability. This allows for smarter matching, including batching orders or delaying assignments to cut costs and improve efficiency. Ignoring these factors increases the risk of order rejection and reliance on costly fallback options \citep{hou2021matching}. Dynamic pricing is equally vital, as fixed payouts fail to reflect changing supply and demand. By adjusting payouts in real time, based on urgency and courier availability, stores can better balance reliability and cost. Several studies in the literature attempt to address this problem from different angles. Some, such as \citet{mousavi2024approximate}, \citet{arslan2019crowdsourced}, and \citet{dayarian2020crowdshipping}, focus primarily on optimizing matching decisions, while either omitting pricing considerations or assuming fixed, non-adaptive compensation schemes. Others, including \citet{le2021designing}, \citet{adil2021dynamic}, and \citet{peng2024outsourcing}, incorporate pricing mechanisms but do so in a static or externally defined manner, without integrating pricing into a unified, forward-looking matching strategy. Moreover, most of these works do not model customer acceptance behavior, limiting their ability to capture real-world decision-making dynamics.

In this paper, we extend the centralized crowd-shipping framework by introducing dynamic pricing mechanisms, building on \citet{mousavi2024approximate}'s work, which model the problem as an MDP and use Approximate Dynamic Programming (ADP) to determine order-to-shopper matching under uncertainty. While their approach yields significant cost savings over myopic baselines, it assumes fixed compensation and guaranteed shopper acceptance, limiting its practical use-cases. Their model also restricts each crowd-shipper to a single delivery outside their home location and uses aggregated delivery zones. We generalize the problem to support multiple, granular delivery destinations, allowing for better route consolidation and a more realistic operational environment. Like \citet{mousavi2024approximate}, we model the problem as an MDP but enrich the state space and jointly optimize matching and pricing. To reflect real-world behavior, we incorporate stochastic acceptance of delivery offers. Building on recent advances in NeurADP and Mean-Field Q-learning for ride-pooling \citep{zhang2023future}, we propose a novel joint optimization approach that combines NeurADP with a DDQN-based pricing policy for centralized same-day crowd-shipping. We evaluate our method against several baselines and conduct a detailed sensitivity analysis to assess robustness across key operational parameters, such as compensation schemes, allowable delivery locations per courier, order-to-shipper ratios, and delivery delay flexibility. Our findings offer data-driven insights for decision-makers implementing dynamic crowd-shipping systems in urban logistics.

The rest of the paper is structured as follows. Section~\ref{CSliteraturereview} reviews the most relevant prior work and situates our study within the existing literature. Section~\ref{CSproblemdescription} introduces the formal problem formulation and outlines the key elements of the centralized crowd-shipping setting under consideration. In Section~\ref{CSsolutionmethodology}, we detail the proposed NeurADP and DDQN-based solution approach, including its integration with dynamic pricing. Section~\ref{CSexperimentalsetup} describes the experimental design, including dataset construction, neural network architectures, and baseline policies used for comparison. Section~\ref{CSresults} reports and analyzes the outcomes of our computational study. Lastly, Section~\ref{CSconclusion} summarizes the main contributions and highlights potential avenues for future investigation.

%%%%%%%%%%%%%%%%%%%%%%%%%%%%%%%%%%%%%%%%%%%%%%%%%%%%%%%%
%%%%%%%%%%%%% Section: Literature Review %%%%%%%%%%%%%%%
%%%%%%%%%%%%%%%%%%%%%%%%%%%%%%%%%%%%%%%%%%%%%%%%%%%%%%%%
\section{Literature Review} \label{CSliteraturereview}

Recent research on crowd-shipping has increasingly emphasized the importance of jointly optimizing pricing and assignment decisions, as platforms seek to balance operational efficiency, crowd-shipper incentives, and customer satisfaction. While many studies explore different aspects of this problem, few address the integrated and forward-looking coordination of pricing and matching under uncertainty, especially in the context of in-store customer couriers.

Several works focus specifically on dynamic pricing within crowd-shipping systems. \citet{le2021designing} jointly optimize sender and courier prices under varying market conditions, showing that individualized pricing can improve platform performance. \citet{adil2021dynamic} and \citet{wu2022dynamic} propose pricing frameworks that adjust wages and service prices based on delivery flexibility, effort levels, and demand fluctuations. \citet{peng2024outsourcing} take a different approach, applying a Stackelberg model to optimize outsourcing prices in a collaborative setup between parcel carriers and on-demand mobility providers. \citet{xiao2023crowd} compare pricing strategies across different platform models, highlighting the central role of price design in ensuring profitability. Empirical studies also contribute to this discussion: \citet{rechavi2022crowd} analyze bidding behaviors using real-world data, while \citet{ermagun2018bid}, \citet{ermagun2020shipment}, and \citet{le2019influencing} estimate price elasticities and willingness to pay, shedding light on how reward amounts and service attributes affect decision-making. 

Beyond pricing, a parallel stream of literature addresses the integration of pricing and matching decisions, though majority of these works focus on ride-sharing rather than crowd-shipping. These studies aim to optimize performance across multiple objectives using a range of modeling techniques. Learning-based approaches such as those by \citet{zhang2023future} and \citet{shah2020neural} apply NeurADP and mean field Q-learning or auction-based pricing to balance immediate revenue with long-term efficiency. \citet{yan2020dynamic} and \citet{ozkan2020joint} propose demand-responsive methods and adjust prices in real time based on spatial demand fluctuations, often paired with graph-based or heuristic matchings. \citet{feng2024two} introduce a two-stage stochastic framework that integrates individualized pricing with predictive assignment, optimizing for both welfare and efficiency. Mechanism design approaches, such as \citet{chen2020pricing} and \citet{hikima2024joint}, frame the joint problem as a min-cost flow or auction, while hybrid systems like \citet{manchella2021passgoodpool} coordinate rule-based pricing with distributed matching for mixed transport modes. 

Reinforcement learning and fairness-aware methods further refine joint policies, as seen in the works by \citet{zhou2023fairness}, \citet{haliem2021distributed}, and \citet{varma2023dynamic}. Complementing these contributions, ADP has proven effective across a wide range of dynamic optimization problems in the relevant domains and tasks. \citet{yang2017approximate} use ADP to price delivery slots in online grocery systems by estimating customer preferences through multinomial logit models. \citet{al2020approximate} apply ADP to manage autonomous vehicle fleets, combining surge pricing with dispatch and recharging decisions to optimize real-time performance. In energy markets, \citet{jiang2015optimal} and \citet{samadi2014real} employ ADP-based methods to dynamically adjust bidding and pricing strategies based on stochastic inputs and system constraints, highlighting the method's adaptability to different domains.

To situate our work within this broader context, we summarize the most relevant studies in Table~\ref{table:CSreferences}. In this table, the column ``Crowd-Shipping'' indicates whether the study addresses a crowd-shipping problem. ``Matching-Optimized'' is checked when the assignment of requests to couriers is formulated as a decision variable in an optimization or learning framework, rather than determined through static rules. ``Pricing-Optimized'' is checked when pricing is learned or jointly optimized through dynamic or data-driven methods, rather than set externally. ``Non-Myopic Matching'' and ``Non-Myopic Pricing'' indicate whether the study incorporates foresight into the matching or pricing decision, respectively, such as reasoning about future supply, demand, or opportunity costs. The final column, ``Sources of Stochasticity,'' identifies whether the system models uncertainty in request arrivals (R), delivery agent arrivals (D), or acceptance behavior (A). Blank entries indicate that the system is fully deterministic or uses static assumptions.

\setlength{\tabcolsep}{2.5pt} % adjust column separation in table
\renewcommand{\arraystretch}{0.95} % adjust row separation in table
\begin{table}[!ht]
\centering
\caption{Summary of the most relevant studies. (R: Requests arrive stochastically, D: Delivery agents arrive stochastically, A: Acceptance of requests is stochastic)}\label{table:CSreferences}
\resizebox{0.99\textwidth}{!}{
% \begin{tabular}{lcccccc}
\begin{tabular}{P{0.29\textwidth}C{0.13\textwidth}C{0.13\textwidth}C{0.13\textwidth}C{0.13\textwidth}C{0.13\textwidth}C{0.13\textwidth}} 
\toprule
\textbf{Study} & \textbf{Crowd-Shipping} & \textbf{Matching Optimized} & \textbf{Pricing Optimized} &  \textbf{Non-Myopic Matching} & \textbf{Non-Myopic Pricing} & \textbf{Sources of Stochasticity}  \\
\midrule

\citet{arslan2019crowdsourced} & \checkmark  & \checkmark &  &  & & [R, D] \\ % In-store crowd-shipping

\citet{dayarian2020same} & \checkmark  & \checkmark &  &  & & [R, D]\\ % In-store crowd-shippers

\citet{haliem2021distributed} &  & \checkmark & \checkmark & \checkmark & \checkmark & [R] \\

\citet{le2021designing} & \checkmark  & \checkmark & \checkmark &  &  &  \\ % No

\citet{adil2021dynamic} & \checkmark  & \checkmark & \checkmark &  &  & [R, D] \\ % In-store crowd-shippers

\citet{shah2022joint} &  & \checkmark  & \checkmark  &  &  & [R, A]  \\

\citet{wu2022dynamic} & \checkmark  &  & \checkmark & & \checkmark & [R]  \\ % No

\citet{zhang2023future} &  & \checkmark  & \checkmark  & \checkmark & \checkmark & [R, A]  \\

\citet{stokkink2024column} & \checkmark  & \checkmark &  & \checkmark &  &  \\ % No

\citet{peng2024outsourcing} & \checkmark  & \checkmark & \checkmark & \checkmark & \checkmark &  \\ % No

\citet{mousavi2024approximate} & \checkmark  & \checkmark &  & \checkmark &  & [R, D] \\ % In-store crowd-shippers

\midrule
\textbf{Our Work} & \checkmark & \checkmark & \checkmark & \checkmark & \checkmark & [R, D, A] \\ 
\bottomrule
\end{tabular}
}
\end{table}

Our work distinguishes itself by simultaneously optimizing matching and pricing decisions in a crowd-shipping context, while explicitly incorporating non-myopic reasoning and modeling multiple sources of uncertainty. Several studies in Table~\ref{table:CSreferences}, such as \citep{adil2021dynamic, arslan2019crowdsourced, dayarian2020same}, also address in-store crowd-shipping with optimized matching, but they either exclude pricing decisions or use purely myopic assignment rules. \citet{le2021designing} and \citet{shah2022joint} jointly optimize pricing and matching but do not account for future demand and supply. A few works, including those by \citet{haliem2021distributed}, \citet{zhang2023future}, and \citet{peng2024outsourcing}, incorporate non-myopic strategies for both components. Of these, \citet{haliem2021distributed} and \citet{zhang2023future} do not address crowd-shipping, while \citet{peng2024outsourcing} analyze a fully deterministic system involving professional drivers rather than in-store customer couriers. None of these works model the level of uncertainty considered in our work. The closest study to our problem setting is \citet{mousavi2024approximate}'s work, which consider in-store crowd-shippers and model order-to-shopper matching in centralized crowd-shipping as an MDP, developing an ADP policy under fixed compensation and guaranteed acceptance assumptions. While their approach outperforms myopic baselines, it limits each courier to a single delivery beyond their home destinations and aggregates customer destinations into coarse zones. We extend this framework by supporting multi-stop routing, modeling probabilistic acceptance, and jointly optimizing both matching and dynamic pricing. Leveraging NeurADP and deep reinforcement learning, we propose a novel hybrid architecture that integrates NeurADP for matching with a DDQN-based pricing policy. These enhancements make our model more reflective of real-world complexities and better suited to support decision-making in dynamic, store-managed crowd-shipping environments.

%%%%%%%%%%%%%%%%%%%%%%%%%%%%%%%%%%%%%%%%%%%%%%%
%%%%%%%%%%%%% Section: Problem Description and Formulation %%%%%%%%%%%%%
%%%%%%%%%%%%%%%%%%%%%%%%%%%%%%%%%%%%%%%%%%%%%%%
\section{Model Description and Formulation} \label{CSproblemdescription}

In this section, we begin by providing a formal definition of our crowd-shipping problem. We then introduce our MDP formulation that serves as the foundation for our proposed solution approach.

\subsection{Problem Definition} 
\label{CSproblemdefinition}

This paper addresses the problem of dynamic matching and pricing in a centralized crowd-shipping system managed by a single brick-and-mortar store, reflecting real-world practices where retailers such as Walmart fulfill online orders directly from the store. In this setup, the store leverages in-store customers, who are already shopping and planning to return home, as potential couriers for delivering incoming online orders. Each order must be delivered within a short time window, defined relative to its arrival time, e.g., within 90 minutes, which reflects customer expectations for timely fulfillment. Any order that remains unserved by its delivery deadline incurs a penalty and is fulfilled through traditional last-mile delivery at the end of the day at a higher cost. At discretized intervals throughout the business day, the store must decide which outstanding orders to offer to eligible in-store shoppers and how much compensation to offer. These offers are either accepted or rejected based on the shopper's willingness, which depends on the payment offered and the extra travel time required compared to driving directly to their originally intended destination. While each online order may contain multiple items, we do not model capacity or weight constraints explicitly. Instead, we limit the number of unique delivery destinations a shopper is willing to visit, which more accurately reflects the operational burden of deviating from their original intended route. Shoppers who are not assigned a delivery at a given decision point are assumed to leave the system immediately, and cannot be reconsidered for future offers. This one-time participation assumption reflects practical behavior, since customers are unlikely to stay in the store beyond the end of their shopping trip just to wait for a possible delivery opportunity.

Given both online orders and in-store shoppers arrive at random times, and acceptance decisions are uncertain, the system must operate under constant evolving conditions that require decisions to be made in the presence of uncertainty. The decision process takes place over a fixed planning horizon, which is divided into discrete time intervals of length~\(\CSTimeIntervals\) (for example, five minutes). Let \(\CSTime := \{0, \ldots, T\}\) represent the set of decision epochs. At each epoch, the system first prices all unmatched orders, including those both new and previously carried-over, using a combination of base compensation and a deviation cost that reflects the extra distance a shopper would need to travel compared to their direct route home. Based on these prices, the system selects assignments and makes delivery offers to the available shoppers. At each time step, orders have the potential to be consolidated, or batched, together to be delivered by a single crowd-shipper. If a shopper accepts an offer, they depart with the assigned order or bundle of orders, and the store incurs the associated cost. If the offer is rejected, the shopper exits the system, and the order (or set of orders) remains available for reassignment in later intervals, provided it can still be delivered on time. Any order that can no longer feasibly be delivered before its deadline is removed from the system and assumed to be fulfilled through a fallback delivery mode at a higher cost. Shoppers who are not offered any assignment also leave the system after the decision point. At the end of each decision interval, the system state is updated to reflect new arrivals of crowd-shippers and orders, accepted and rejected offers, and the removal of infeasible orders.

This model reflects a growing interest in leveraging the idle capacity of in-store customers for last-mile delivery, offering cost-efficient fulfillment without relying on dedicated courier fleets. Such a model enables brick-and-mortar retailers to enhance the flexibility of their delivery services while reducing costs and maintaining service-level guarantees. The primary objective of our model is to minimize the total operational cost incurred over the course of the business day. This cost includes payments to crowd-shippers as well as penalties for unserved deliveries. To achieve this, our assignment and pricing decisions consider future order and customer arrival uncertainties, as well as acceptance and rejection possibilities, and the potential downstream impact of current decisions. To handle the complexity of these decisions, we formulate an MDP model and adapt a joint NeurADP and DDQN matching and pricing strategy, enabling effective real-time decision-making under uncertainty. To this end, we partition the finite planning horizon into discrete intervals of duration \(\CSTimeIntervals\), with decisions made at the start of each interval. New orders and shopper arrivals are observed continuously, and the system state is updated after each interval based on the decisions and observed information.

\begin{figure}[!ht]
\begin{center}
\includegraphics[width=0.99\textwidth]{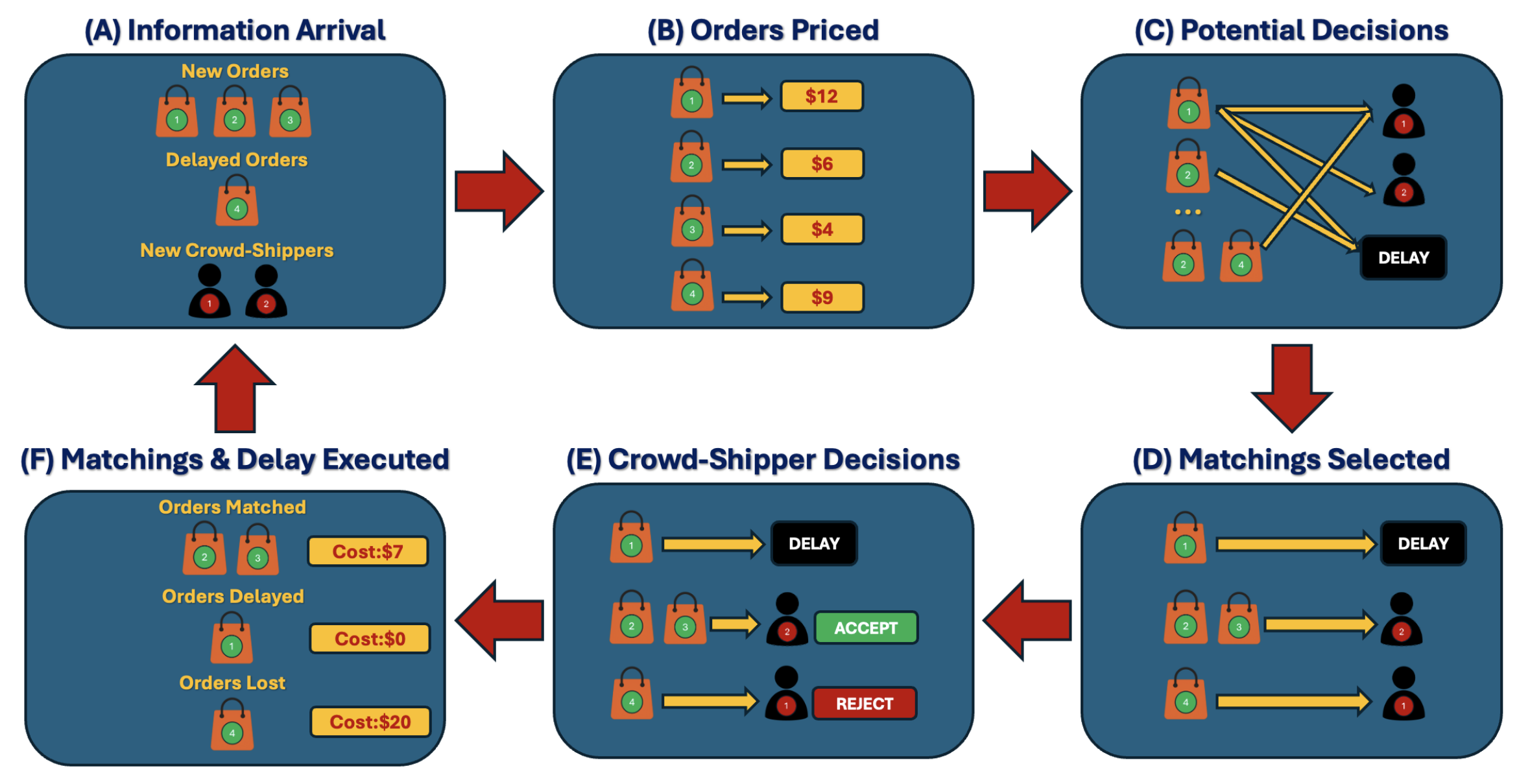}
\end{center}
\caption{Illustrative example of problem dynamics.}
\label{fig:CSproblem_dynamics}
\end{figure}

Figure~\ref{fig:CSproblem_dynamics} illustrates the step-by-step dynamics of the decision-making process in the proposed crowd-shipping framework. In (A), the system observes new online orders, delayed orders carried over from previous intervals, and newly available in-store shoppers. In (B), each order is priced. In (C), all feasible matchings between orders and shoppers are considered, along with the option to delay unassigned orders. In (D), a subset of these matchings is selected for execution based on the systems pricing and assignment strategy. In (E), shoppers make decisions to accept or reject their assigned deliveries. Accepted orders are dispatched, rejected ones remain unmatched, and unassigned shoppers exit the system. Finally, in (F), the outcomes are realized: matched orders incur delivery costs, delayed orders remain in the system, and expired orders incur higher fallback fulfillment costs. This process repeats at each decision epoch until the end of the problem horizon.

% %%%%%%%%%%%%%%%%%
% %%%%%%%%%%%%%%%%%

\subsection{MDP Formulation} 
\label{CSmdp}

In the following subsection, we define the MDP model for our crowd-shipping problem. We begin by defining the state and decision variables, followed by a description of the exogenous information that influences the system. We then present the transition function that governs the evolution of the system over time. Finally, we specify the objective function that guides decision-making within the MDP framework.

\subsubsection{State Variables}

The system state at time \(\CSCurrentTime \in \CSTime\) is defined as \(\CSStateNote_\CSCurrentTime = (\CSCrowdShipperState_\CSCurrentTime, \CSRequestState_\CSCurrentTime)\), where \(\CSCrowdShipperState_\CSCurrentTime\) denotes the state of active crowd-shippers and \(\CSRequestState_\CSCurrentTime\) represents the state of outstanding order requests awaiting delivery. Each crowd-shipper \(\CSIndividualCrowdShipper \in \CSCrowdShipperSet\), where \(\CSCrowdShipperSet\) represents the set of all possible crowd-shipper states, is associated with a home destination \(\CSIndividualCrowdShipper_\CSCrowdShipperHome\), which denotes the location they are expected to return to after completing their deliveries. Each order request \(\CSIndividualOrderRequest \in \CSOrderRequestSet\), where \(\CSOrderRequestSet\) represents the set of all order request states, is defined by two key attributes: the delivery destination \(\CSIndividualOrderRequest_\CSRequestDestination\), and the number of decision epochs remaining before the order exits the crowd-shipping system, denoted by \(\CSIndividualOrderRequest_\CSRequestEpochsRemaining\). Additionally, \(\CSIndividualOrderRequest_\CSRequestDeadline\) represents the exact time by which the order must be delivered. While \(\CSIndividualOrderRequest_\CSRequestEpochsRemaining\) and \(\CSIndividualOrderRequest_\CSRequestDeadline\) convey the same underlying delivery constraint, we define both for clarity and notational convenience in subsequent model components. The value of \(\CSIndividualOrderRequest_\CSRequestEpochsRemaining\) is initialized when the order enters the system and reflects the number of future decision epochs during which the order can still be matched and feasibly delivered. If \(\CSIndividualOrderRequest_\CSRequestEpochsRemaining = 0\), the order must be matched in the current epoch or it will exit the system unfulfilled. This value is determined by the order's delivery deadline, the time it entered the system, and the direct travel time required to serve it from the store. Let \(\CSLocations = \{0, 1, \ldots, L\}\) denote the locations in the system, where \(\ell = 0\) denotes the store, and the rest represent order destinations and shopper home locations. The travel time from location \(\ell\) to \(\ell'\), departing from \(\ell\), is denoted by \(\CSTravelTime(\ell, \ell')\). Let \(\CSDeadlineTime\) denote the maximum allowable delay (e.g., \(\CSDeadlineTime = 90\) implies a 90-minute window for delivery after the order enters the system), and let \(\CSTravelTime(0, \CSIndividualOrderRequest_\CSRequestDestination)\) represent the direct travel time from the store to the order's destination. We assume that every order entering the system is directly deliverable within the deadline, i.e., \(\CSTravelTime(0, \CSIndividualOrderRequest_\CSRequestDestination) \leq \CSDeadlineTime\). Given that \(\CSEpochLength\) is the length of a single decision epoch, the initial number of feasible epochs is calculated as:
\begin{equation}
\CSIndividualOrderRequest_\CSRequestEpochsRemaining = \left\lfloor \frac{\CSDeadlineTime - \CSTravelTime(0, \CSIndividualOrderRequest_\CSRequestDestination)}{\CSEpochLength} \right\rfloor
\label{eq:CSremaining_epochs}
\end{equation}
This expression captures the number of decision epochs remaining before the order must be matched in order to be delivered on time, assuming it is sent directly from the store to the destination without any intermediate stops. After each decision epoch, this value is reduced by one. Building on the state representation, we next define the decision variables which guide pricing and matching decisions in our model.

% At each decision epoch $\CSCurrentTime \in \CSTime$, the central store observes the current system state $\CSStateNote_\CSCurrentTime$ and selects a set of actions from the feasible action set $\CSAllActionsSet_\CSCurrentTime(\CSStateNote_\CSCurrentTime)$. The decision variables are defined as follows:

% \begin{itemize}
%     \item $\CSPricingVector$: A vector of pricing multipliers $\CSPricingIndividualMultiplier_{\CSIndividualOrderRequest} \in \CSPricingDecisionSet$ for each order $\CSIndividualOrderRequest \in \CSOrderRequestSet_\CSCurrentTime$, used to determine the offered compensation to crowd-shippers.
    
%     \item $\CSActionVector$: A set of binary assignment variables $\CSActionVar_{\CSCurrentTime \CSSingleBatch \CSIndividualCrowdShipper} \in \{0,1\}$ indicating whether a batch of orders $\CSSingleBatch$ is assigned to a crowd-shipper $\CSIndividualCrowdShipper$.
    
%     \item $\CSDelayVector$: A set of binary variables $\CSDelayVar_{\CSCurrentTime \CSIndividualOrderRequest} \in \{0,1\}$ indicating whether an order $\CSIndividualOrderRequest$ is delayed to the next epoch.
% \end{itemize}

% Together, these define the action vector $\CSDecisionsVector = (\CSPricingVector, \CSActionVector, \CSDelayVector)$, which governs the store's decisions at each time step.

%%%%%%%%%%%%%%%

\subsubsection{Decision Variables}

At each decision epoch $\CSCurrentTime \in \CSTime$, the central store makes three types of decisions: setting prices for individual orders, assigning orders (either individually or in batches) to crowd-shippers, and choosing whether to delay orders. We begin by defining the feasibility conditions for each of these actions and then introduce the overall decision vector. To begin, pricing decisions are made individually for each order. Every order is assigned a price multiplier, $\CSPricingIndividualMultiplier$, chosen from a predefined discrete set, $\CSPricingDecisionSet$, which adjusts the delivery compensation offered to the crowd-shipper. These multipliers are designed to influence the likelihood that a crowd-shipper will accept the delivery offer. Once pricing is complete, the store constructs candidate batches of orders. Collection of all possible batches of orders can be expressed as 
\begin{align}
\CSPossibleBatchesSet \coloneqq \left\{ \CSSingleBatch \subseteq \CSOrderRequestSet \;\middle|\; |\CSSingleBatch| \leq \CSMaxLocations \right\}
\end{align} 
Wherein $\CSPossibleBatchesSet$ is a subset of the power set \(2^{\CSOrderRequestSet}\), such that each batch \(\CSSingleBatch \in \CSPossibleBatchesSet\) contains no more than \(\CSMaxLocations\) orders. Here, \(\CSMaxLocations\) denotes the maximum number of delivery locations a crowd-shipper is willing to visit in a single trip (not including their home location). For a batch $\CSSingleBatch = \{\CSIndividualOrderRequest_1, \dots, \CSIndividualOrderRequest_n\}$ and crowd-shipper $\CSIndividualCrowdShipper$, let $\CSPossibleRoutesSet_{\CSSingleBatch\CSIndividualCrowdShipper}$ denote the set of all potential delivery routes. Each route begins at the store (denoted by 0), visits all delivery destinations in a given sequence, and ends at the crowd-shipper's home location. Formally:
\begin{equation}
\CSPossibleRoutesSet_{\CSSingleBatch \CSIndividualCrowdShipper} = 
\left\{ 
\CSRoute = \left(0, \CSIndividualOrderRequest_{\CSOrderPermutation(1),\CSRequestDestination}, \dots, \CSIndividualOrderRequest_{\CSOrderPermutation(n),\CSRequestDestination}, \CSIndividualCrowdShipper_\CSCrowdShipperHome \right)
\;\middle|\;
\CSOrderPermutation \in \CSPermutationSet(\CSSingleBatch)
\right\}.
\end{equation}
Here, $\CSPermutationSet(\CSSingleBatch)$ denotes the set of all permutations over the orders in $\CSSingleBatch$, applied only to the delivery destinations. The store location is always the starting location, and the crowd-shipper's home $\CSIndividualCrowdShipper_\CSCrowdShipperHome$ is always the final stop in every route. 

A route $\CSRoute \in \CSPossibleRoutesSet_{\CSSingleBatch \CSIndividualCrowdShipper}$ is considered feasible if all orders can be delivered before their respective deadlines. Let $\CSIndividualOrderRequest_{\CSRequestDeadline}$ denote the delivery deadline associated with order $\CSIndividualOrderRequest$. We impose the following constraint:
\begin{equation}
\CSCurrentTime
  + \sum_{k = 1}^{j} 
    \CSTravelTime\!\left(\CSRoute_{k}, \CSRoute_{k + 1}\right)
  \,\leq\,
  \CSIndividualOrderRequest_{\CSOrderPermutation(j), \CSRequestDeadline},
\qquad
j = 1, \dots, n,
\label{eq:CSdeadline-constraint}
\end{equation}
where $\CSRoute_k$ denotes the $k^{\text{th}}$ location in the route, and $\CSCurrentTime$ represents the time at which the batch is considered to be matched and dispatched from the store. The cumulative sum captures the elapsed travel time to reach the $j^{\text{th}}$ delivery location in the route. We adopt a 0-based indexing convention for the route vector $\CSRoute$, and we apply this consistently throughout the paper. Specifically, $\CSRoute_0 = 0$ represents the store, $\CSRoute_1, \dots, \CSRoute_n$ are the delivery destinations, and $\CSRoute_{n+1} = \CSIndividualCrowdShipper_\CSCrowdShipperHome$ is the home location of the crowd-shipper. Let $\CSTotalTravelTime(\CSRoute)$ denote the total travel time along route $\CSRoute \in \CSPossibleRoutesSet_{\CSSingleBatch \CSIndividualCrowdShipper}$, defined as:
\begin{equation}
\CSTotalTravelTime(\CSRoute) = \sum_{k=1}^{n+1} \CSTravelTime(\CSRoute_{k-1}, \CSRoute_k).
\label{eq:CStotal-travel-time}
\end{equation}
Among all feasible delivery routes that satisfy Constraint~\eqref{eq:CSdeadline-constraint}, the optimal route $\CSBestRoute_{\CSSingleBatch\CSIndividualCrowdShipper}$ is selected as:
\begin{equation}
\CSBestRoute_{\CSSingleBatch\CSIndividualCrowdShipper} \in \arg\min_{\CSRoute \in \CSPossibleRoutesSet_{\CSSingleBatch \CSIndividualCrowdShipper}} \left\{ \CSTotalTravelTime(\CSRoute) \;\middle|\; \eqref{eq:CSdeadline-constraint} \right\}.
\label{eq:CSoptimal-route}
\end{equation}
This ensures that the selected route not only satisfies all delivery deadlines but also minimizes the total delivery and return travel time, thereby reducing the detour burden imposed on the crowd-shipper. Thus, a batch $\CSSingleBatch$ is deemed feasible for crowd-shipper $\CSIndividualCrowdShipper$ if there exists at least one such route $\CSRoute$ that satisfies the constraint in \eqref{eq:CSdeadline-constraint}. The overall set of feasible matching decisions at time $\CSCurrentTime$ is given by:
\begin{equation}
\CSFeasibleMatchingSet_\CSCurrentTime = \left\{ 
(\CSSingleBatch, \CSIndividualCrowdShipper) \;\middle|\;
\CSSingleBatch \in \CSPossibleBatchesSet_\CSCurrentTime, 
\exists \CSRoute \in \CSPossibleRoutesSet_{\CSSingleBatch \CSIndividualCrowdShipper} 
\text{ s.t. } \eqref{eq:CSdeadline-constraint}
\right\}.
\end{equation}
Finally, in addition to assigning order batches to crowd-shippers, the store may also choose to delay individual orders. A delay action can be taken on any order currently present in the system, i.e., any $\CSIndividualOrderRequest \in \CSOrderRequestSet$. This includes orders with $\CSIndividualOrderRequest_{\CSRequestEpochsRemaining} = 0$, which will exit the system before the next decision epoch but are still eligible for a delay action at the current time step. Thus, we denote the overall decision at time $\CSCurrentTime$ as $\CSDecisionsVector = (\CSPricingVector, \CSActionVector, \CSDelayVector) \in \CSAllActionsSet_\CSCurrentTime(\CSStateNote_\CSCurrentTime)$, where $\CSPricingVector$ contains the pricing choices for each order, $\CSActionVector$ captures the assignment of order batches to crowd-shippers, and $\CSDelayVector$ specifies which orders are delayed.

\subsubsection{Cost Evaluation}

At each decision epoch $\CSCurrentTime \in \CSTime$, the store must determine whether to delay an order $\CSIndividualOrderRequest$ to the next time-step or assign it to a crowd-shipper as part of a batch $\CSSingleBatch$. This decision involves evaluating the cost associated with each option. The cost of delaying an order, denoted as $\CSDelayCost_{\CSIndividualOrderRequest}$, depends on the order's remaining delivery time. If the order has no epochs remaining before its deadline (i.e., $\CSIndividualOrderRequest_\CSRequestEpochsRemaining = 0$) when delayed, it must be fulfilled through an external fallback mechanism, incurring a fixed penalty $\CSLostCost$. Otherwise, no immediate cost is applied. This is formally defined as:
\begin{equation}
\CSDelayCost_{\CSIndividualOrderRequest} =
\begin{cases}
\CSLostCost, & \text{if } \CSIndividualOrderRequest_\CSRequestEpochsRemaining = 0 \\
0, & \text{otherwise}.
\end{cases}
\label{eq:CSdelay-cost-order}
\end{equation}

Subsequently, if a batch of orders is offered to a crowd-shipper, given that crowd-shippers may accept or reject the offering, there are two potential associated costs incurred. First, if the crowd-shipper rejects the offer, then the cost incurred is the sum of the delay costs for the orders associated with the batch $\CSSingleBatch$, formally defined as 
\begin{equation}
\CSDelayCost_{\CSSingleBatch} = \sum_{\CSIndividualOrderRequest \in \CSSingleBatch}
\CSDelayCost_{\CSIndividualOrderRequest}
\label{eq:CSdelay-cost-batch}
\end{equation}

In the second case, if the batch of orders is accepted for delivery by the crowd-shipper, the store incurs a direct matching cost consisting of two components: (i) the total adjusted compensation for the orders in the batch, and (ii) a detour penalty that accounts for the additional effort required from the crowd-shipper to complete the deliveries. The first component is derived from a base compensation fee, denoted $\CSBaseCompensation$, which applies to all assigned orders. This fee is scaled by a pricing multiplier selected by the store for each individual order. The adjusted compensation for an order $\CSIndividualOrderRequest$ is formally defined as follows:
\begin{equation}
\CSAdjustedBaseFee_\CSIndividualOrderRequest = \CSPricingIndividualMultiplier_{\CSIndividualOrderRequest} \cdot \CSBaseCompensation,
\label{eq:CSadjusted-base-fee}
\end{equation}
wherein $\CSPricingIndividualMultiplier_\CSIndividualOrderRequest$ is the pricing multiplier applied to order $\CSIndividualOrderRequest$. We then define the total adjusted and base compensation for a batch $\CSSingleBatch$ as
\begin{equation}
\CSAdjustedBaseFee_\CSSingleBatch = \sum_{\CSIndividualOrderRequest \in \CSSingleBatch} \CSAdjustedBaseFee_\CSIndividualOrderRequest, \quad
\CSBaseCompensation_\CSSingleBatch = \sum_{\CSIndividualOrderRequest \in \CSSingleBatch} \CSBaseCompensation.
\label{eq:CSbatch-compensation}
\end{equation}
These multipliers allow the store to dynamically adjust the perceived value of each order in real time, based on factors such as urgency or delivery distance. For example, an order nearing its delivery deadline or with a remote destination may receive a higher multiplier to increase its appeal to crowd-shippers. In contrast, orders with more slack time or closer drop-off locations may be priced more conservatively. The second component is a detour or deviation fee, denoted by $\CSDetourCompensation$, which represents the per-minute compensation the store pays for each minute the crowd-shipper spends deviating from their original route to their intended destination, such as their home. This fee is multiplied by the total additional time required to complete all assigned deliveries. For instance, if the direct route home from the store takes 10 minutes for the crowd-shipper, and the delivery route takes them 15 minutes, the detour time is 5 minutes. This term quantifies the extra effort imposed on the crowd-shipper by comparing the delivery route duration to the direct travel time home. 

We formally define the direct matching cost as follows:
\begin{equation}
\CSMatchCost_{\CSSingleBatch \CSIndividualCrowdShipper} = \CSAdjustedBaseFee_\CSSingleBatch + \CSDetourTime_{\CSSingleBatch \CSIndividualCrowdShipper} \cdot \CSDetourCompensation,
\label{eq:CSmatching-cost}
\end{equation}
where the detour time is defined as
\begin{equation}
\CSDetourTime_{\CSSingleBatch \CSIndividualCrowdShipper} = \CSTotalTravelTime(\CSBestRoute_{\CSSingleBatch \CSIndividualCrowdShipper}) - \CSTravelTime(0,\CSIndividualCrowdShipper_\CSCrowdShipperHome).
\label{eq:CSdetour-time}
\end{equation}
Thus, the expected cost of matching a batch of orders $\CSSingleBatch$ to a crowd-shipper $\CSIndividualCrowdShipper$, given the probabilistic nature of acceptance, is as follows:
\begin{equation}
\CSExpectedCost_{\CSSingleBatch \CSIndividualCrowdShipper} = \CSAcceptProbFunc(\CSSingleBatch) \cdot \CSMatchCost_{\CSSingleBatch \CSIndividualCrowdShipper} + \left(1 - \CSAcceptProbFunc(\CSSingleBatch) \right) \cdot \CSDelayCost_\CSSingleBatch,
\label{eq:CSexpected-match-cost}
\end{equation}
wherein $\CSAcceptProbFunc$ denotes the acceptance probability function and is defined as
\begin{equation}
\CSAcceptProbFunc(\CSSingleBatch) = \frac{1}{1 + e^{5 \cdot \left( \frac{\CSBaseCompensation_\CSSingleBatch}{\CSAdjustedBaseFee_\CSSingleBatch} \right) - 5.5}}
\label{eq:CSaccept-prob}
\end{equation}

The acceptance function in (\ref{eq:CSaccept-prob}) is adapted from prior work in the ride-pooling literature \citep{shah2022joint, zhang2023future}, where similar models are used to capture user decision-making. In our setting, the function is modified to better reflect the dynamics of crowd-shipping, where acceptance decisions are highly sensitive to perceived compensation. The formulation captures the idea that crowd-shippers are much more likely to accept a delivery when the reward shows a clear improvement over the baseline. The steep, nonlinear shape of the function emphasizes this behavior by sharply increasing acceptance probability when the offer exceeds expectations, and quickly dropping off when it does not. This structure is particularly suited to our setting, where participants are short-term, in-store shoppers deciding whether it is worth the extra effort to take on a delivery.

The cost incurred at time $\CSCurrentTime$ is given by
\begin{equation}
    \CSCostName_\CSCurrentTime(\CSDecisionsVector) = 
    \sum_{(\CSSingleBatch, \CSIndividualCrowdShipper) \in \CSFeasibleMatchingSet_\CSCurrentTime} 
    \CSActionVar_{\CSCurrentTime \CSSingleBatch \CSIndividualCrowdShipper} \cdot \CSExpectedCost_{\CSSingleBatch \CSIndividualCrowdShipper} 
    + \sum_{\CSIndividualOrderRequest \in \CSOrderRequestSet_\CSCurrentTime} 
    \CSDelayVar_{\CSCurrentTime \CSIndividualOrderRequest} \cdot \CSDelayCost_{\CSIndividualOrderRequest},
\label{eq:CScost-timestep}
\end{equation}
where $\CSActionVar_{\CSCurrentTime \CSSingleBatch \CSIndividualCrowdShipper} \in \{0,1\}$ indicates whether batch $\CSSingleBatch$ is assigned to crowd-shipper $\CSIndividualCrowdShipper$ at time $\CSCurrentTime$, with 1 representing an assignment and 0 otherwise. Similarly, $\CSDelayVar_{\CSCurrentTime \CSIndividualOrderRequest} \in \{0,1\}$ denotes whether order $\CSIndividualOrderRequest$ is delayed at time $\CSCurrentTime$. The first term captures the expected cost of assigning batches to crowd-shippers, which may be influenced by earlier pricing decisions. The second term accounts for the penalty incurred by delaying unassigned orders. Together, these terms represent the trade-off between fulfilling orders immediately and deferring them to future decision epochs.

\subsubsection{Random Information}
\label{CSexogenous}

At the end of epoch $\CSCurrentTime$ we draw the random vector
\begin{equation}
\CSExogenousInformation_{\CSCurrentTime+1}
= \left( \CSExogenousInformation^{\CSAcc}_{\CSCurrentTime+1},
         \CSExogenousInformation^{\CSCrowd}_{\CSCurrentTime+1},
         \CSExogenousInformation^{\CSOrd}_{\CSCurrentTime+1} \right).
\end{equation}
The first component, $\CSExogenousInformation^{\CSAcc}_{\CSCurrentTime+1}$, records whether each batch that was just priced and offered is accepted. For every batch $\CSSingleBatch$ there is a Bernoulli variable $\CSExogenousInformation^{\CSAcc}_{\CSCurrentTime+1, \CSSingleBatch}$ that takes the value 1 when the shopper accepts and 0 when the shopper rejects, and the success probability is the acceptance function $\CSAcceptProbFunc(\CSSingleBatch)$. Because that probability depends on the prices chosen in the decision vector $\CSDecisionsVector$, the distribution of $\CSExogenousInformation^{\CSAcc}_{\CSCurrentTime+1}$ is endogenous to the action, though its realized value remains beyond the store's control once the decision has been taken. The second component, $\CSExogenousInformation^{\CSCrowd}_{\CSCurrentTime+1}$, represents new incoming shoppers who complete their in-store activities during the interval $(\CSCurrentTime, \CSCurrentTime + 1]$ and are about to head home. This information is considered purely exogenous, as it is determined independently of the store's pricing or matching decisions. The third component, $\CSExogenousInformation^{\CSOrd}_{\CSCurrentTime+1}$, holds the destinations and deadlines of all online orders that arrive in the same interval. Like the shopper arrivals, its information is action-independent and therefore exogenous.
We define $\CSPricingDistribution_{\CSCurrentTime+1 | \CSPricingVector}$ as the conditional distribution over $\CSExogenousInformation_{\CSCurrentTime+1}$ given the pricing decision vector $\CSPricingVector$. This distribution captures the endogenous nature of the acceptance outcomes, where each $\CSExogenousInformation^{\CSAcc}_{\CSCurrentTime+1, \CSSingleBatch}$ follows a Bernoulli distribution with success probability $\CSAcceptProbFunc(\CSSingleBatch)$. The remaining components, $\CSExogenousInformation^{\CSCrowd}_{\CSCurrentTime+1}$ and $\CSExogenousInformation^{\CSOrd}_{\CSCurrentTime+1}$, are modeled as exogenous and independent of the pricing decisions.

% Conditional on the chosen decision vector $\CSDecisionsVector$ the three components are mutually independent. This separation lets us treat the post-decision state as deterministic, with all randomness entering only through $\CSExogenousInformation_{\CSCurrentTime+1}$ when the system advances to epoch $\CSCurrentTime + 1$.

\subsubsection{Transition Function}
\label{CStransition}

The transition function from state $\CSStateNote_{\CSCurrentTime}$ to $\CSStateNote_{\CSCurrentTime+1}$, wherein $\CSStateNote_{\CSCurrentTime+1} = (\CSCrowdShipperState_{\CSCurrentTime + 1}, \CSRequestState_{\CSCurrentTime + 1})$, depends on the decision vector $\CSDecisionsVector$ and the realization of the random information vector $\CSExogenousInformation_{\CSCurrentTime+1}$. By adopting the concept of the post-decision state (see \citet{powell2007approximate}), which refers to the state immediately after an action is taken but before new information is revealed, we can decompose the transition into two steps: (i) the transition from the current state to the post-decision state, $\CSPostDecisionState_{\CSCurrentTime}$, through action $\CSDecisionsVector$, and (ii) the transition from the post-decision state to the next state based on the realization of $\CSExogenousInformation_{\CSCurrentTime+1}$.
\begin{equation}
\CSPostDecisionState_{\CSCurrentTime}
      =\CSpost \bigl(
          \CSStateNote_{\CSCurrentTime},\,
          \CSDecisionsVector
       \bigr),\qquad
\CSStateNote_{\CSCurrentTime+1}
      =\CSnext\bigl(
          \CSPostDecisionState_{\CSCurrentTime},\,
          \CSExogenousInformation_{\CSCurrentTime+1}
       \bigr).
\label{eq:CSstep-decomp}
\end{equation}

Given that every in-store crowd-shipper either departs with an assigned batch or leaves empty-handed as soon as the offers are issued, the post-decision state contains no crowd-shippers, only orders:
\begin{equation}
    \CSPostDecisionState_\CSCurrentTime = (\emptyset,\, \CSOrderPost_\CSCurrentTime).
\label{eq:CSpost-dec}
\end{equation}
The set $\CSOrderPost_\CSCurrentTime$ consists of all orders from $\CSStateNote_{\CSCurrentTime}$, moved forward in time and updated to reflect the impact of the current decisions at time $\CSCurrentTime$. For each of these orders, the field $\CSIndividualOrderRequest_\CSRequestEpochsRemaining$ is reduced by one. In the second part of the transition, \CSnext, the realization of $\CSExogenousInformation_{\CSCurrentTime+1}$ reveals which offered orders were accepted or rejected by crowd-shippers, as well as the incoming orders and crowd-shippers. Orders that were offered but rejected in the previous time-step remain in the system, as do orders that were not offered at all, meaning those assigned a delay action, provided they still have a positive value for $\CSIndividualOrderRequest_\CSRequestEpochsRemaining$. Orders are removed from the system if they fall into either of two categories: those that were matched and accepted by a crowd-shipper, or those that were not matched and had $\CSIndividualOrderRequest_\CSRequestEpochsRemaining = 0$ at the current time step. The remaining orders, which were either delayed or rejected and still have time left to be matched, are denoted by $\CSOrderRetained_{\CSCurrentTime}$. These orders are then combined with the new orders arriving as part of $\CSExogenousInformation^{\CSOrd}_{\CSCurrentTime+1}$ to form the order component of the next state as follows:
\begin{equation}
\CSRequestState_{\CSCurrentTime+1} = \CSOrderRetained_{\CSCurrentTime} \cup \CSExogenousInformation^{\CSOrd}_{\CSCurrentTime+1}
\label{eq:CSnew-order}
\end{equation}

Since customers who express willingness to crowd-ship are assumed to leave shortly after doing so, only newly arriving crowd-shippers are included in the next state, that is,
\begin{equation}
\CSCrowdShipperState_{\CSCurrentTime+1} = \CSExogenousInformation^{\CSCrowd}_{\CSCurrentTime+1}
\label{eq:CSnew-crowd}
\end{equation}
The full system state at time $\CSCurrentTime + 1$ is then given by the tuple $\CSStateNote_{\CSCurrentTime + 1} = (\CSCrowdShipperState_{\CSCurrentTime + 1}, \CSRequestState_{\CSCurrentTime + 1})$.

\subsubsection{Objective Function}
\label{sec:CSobj-section}

The objective in the crowd-shipping problem is to minimize the total expected operational cost over the planning horizon, where costs arise from delaying orders, offering delivery batches to shoppers, and potential penalties for unfulfilled deliveries. Because the shopper acceptance outcomes depend on the pricing decisions taken by the policy, the distribution of future uncertainty is policy-dependent. Formally, the objective is expressed as:
\begin{equation}
    \min_{\CSSmallPolicy \in \CSBigPolicy}
    \mathbb{E}_{\CSExogenousInformation=(\CSExogenousInformation_0, \ldots, \CSExogenousInformation_{T}) \sim \CSPricingDistribution^\CSSmallPolicy}
    \left[
        \sum_{\CSCurrentTime \in \CSTime}
        \CSCostName_\CSCurrentTime
        \left( \CSAllActionsSet_\CSCurrentTime^\CSSmallPolicy(\CSStateNote_\CSCurrentTime^\CSSmallPolicy(\CSExogenousInformation)) \right)
        \,\big|\, \CSStateNote_0
    \right].
\label{eq:CSNeurADPObj}
\end{equation}
By solving Equation~\eqref{eq:CSNeurADPObj}, we seek a policy $\CSSmallPolicy \in \CSBigPolicy$ that minimizes the expected cumulative cost, accounting for the endogenous influence of pricing decisions on shopper acceptance. The expectation is taken with respect to $\CSPricingDistribution^\CSSmallPolicy$, the joint distribution over the sequence of information vectors $(\CSExogenousInformation_0, \dots, \CSExogenousInformation_T)$ encountered under policy~$\CSSmallPolicy$. This distribution captures both the endogenous acceptance outcomes, which depend on pricing decisions selected by the policy, and the exogenous components, i.e., shopper arrivals and online order arrivals, which evolve independently of the policy. The realized states evolve according to the following recursion:
\begin{subequations}
\label{m:CSTransitions2}
    \begin{alignat}{2}
    & \CSStateNote_0^\CSSmallPolicy(\CSExogenousInformation)=\CSStateNote_0  \label{eq:CSOptOne} \\
    & \CSStateNote_{\CSCurrentTime+1}^\CSSmallPolicy(\CSExogenousInformation)=\CSnext\left(\CSpost\left(\CSStateNote_\CSCurrentTime^\CSSmallPolicy(\CSExogenousInformation), \CSAllActionsSet_\CSCurrentTime^\CSSmallPolicy(\CSStateNote_\CSCurrentTime^\CSSmallPolicy(\CSExogenousInformation))\right),\CSExogenousInformation_{\CSCurrentTime+1}\right), \quad \CSCurrentTime=0,\ldots,T-1  \label{eq:CSOptTwo}
    \end{alignat}
\end{subequations}
Here, $\CSStateNote_0$ encodes the initial state of the system, including the initial set of orders and available crowd-shoppers at the start of the planning horizon. The vector $\CSExogenousInformation$ represents the full stochastic trajectory of the system, consisting of shopper arrivals, order arrivals, and acceptance outcomes. The expectation in~\eqref{eq:CSNeurADPObj} is taken with respect to the policy-dependent distribution $\CSPricingDistribution^\CSSmallPolicy$, which captures both the exogenous arrival processes and the endogenous acceptance outcomes governed by the pricing decisions selected by policy~$\CSSmallPolicy$. At each decision epoch, the actions are selected based on the realized history up to that point, and influence both the post-decision state and the distribution of future outcomes. As a result, the realized state trajectory depends on both the policy and the unfolding randomness over the horizon. To characterize the optimal value at each state $\CSStateNote_\CSCurrentTime$, we rely on the Bellman optimality equation:
\begin{equation}
\label{eq:CSBellman}
\begin{aligned}
\CSValueNote_\CSCurrentTime(\CSStateNote_\CSCurrentTime) =
\min_{\CSPricingVector} \,
\mathbb{E}_{\CSExogenousInformation_{\CSCurrentTime+1} \sim \CSPricingDistribution_{\CSCurrentTime+1|\CSPricingVector}} \Biggl[
  \min_{(\CSActionVector, \CSDelayVector) \in \CSAllActionsSet_\CSCurrentTime(\CSStateNote_\CSCurrentTime, \CSExogenousInformation_{\CSCurrentTime+1})}
  \Bigl\{
    \CSCostName_\CSCurrentTime(\CSDecisionsVector)
    + \mathbb{E} \left[
      \CSValueNote_{\CSCurrentTime+1}(\CSStateNote_{\CSCurrentTime+1})
    \right]
  \Bigr\}
\Biggr]
\end{aligned}
\end{equation}
where the next state is given by $\CSStateNote_{\CSCurrentTime + 1} = \CSnext\left(\CSpost(\CSStateNote_\CSCurrentTime,\CSDecisionsVector),\CSExogenousInformation_{\CSCurrentTime+1}\right)$, such that the realization of $\CSExogenousInformation_{\CSCurrentTime+1}$ depends on both exogenous arrivals and the endogenous acceptance outcomes induced by pricing decisions. The value function $\CSValueNote_\CSCurrentTime(\cdot)$ represents the optimal expected downstream cost from state $\CSStateNote_\CSCurrentTime$ onward and is defined recursively through the Bellman equation. Although Equation~\eqref{eq:CSBellman} admits an exact solution in principle via dynamic programming, computing it explicitly is intractable in practice. At each epoch, the store must assign prices to all outstanding orders and subsequently select a combinatorial joint matching–delay decision, potentially forming many batches and evaluating each against every available crowd-shipper. The number of admissible pricing vectors grows exponentially with the number of orders, while the size of the matching space expands combinatorially with the number of orders and crowd-shippers. Moreover, because each action influences not just the immediate cost but also the distribution over future outcomes, due to both stochastic arrivals and price-sensitive acceptance, the resulting transition dynamics cannot be pre-enumerated or compactly expressed. This explosion of the state, action, and information spaces motivates the use of reinforcement learning and ADP techniques. In the following section, we adopt a hybrid approach: a Double Deep Q-Network (DDQN) component that learns to adjust prices, combined with a NeurADP value function approximation that guides the combinatorial matching decisions. Together, these components derive a policy that minimizes long-run operational costs under uncertainty.

%%%%%%%%%%%%%%%%%%%%%%%%%%%%%%%%%%%%%%%%%
%%%% Section: Solution Methodology %%%%
%%%%%%%%%%%%%%%%%%%%%%%%%%%%%%%%%%%%%%%%%
\section{Solution Methodology} \label{CSsolutionmethodology}

The crowd-shipping platform considered in this paper involves a sequence of tightly connected pricing and matching decisions, each with downstream consequences. Pricing choices influence the likelihood that crowd-shippers accept delivery offers and shape the perceived attractiveness of each order. These decisions inherently determine which orders are accepted immediately and which remain in the system. In turn, the feasibility and cost of matching depend not only on the current pool of orders and shoppers, but also on how prior decisions have affected system congestion and the spatial distribution of unfulfilled orders. Because deadlines are short and arrivals are uncertain, poorly priced or delayed orders can quickly reduce future matching flexibility and increase the reliance on costly fallback deliveries. Prematurely assigning a batch may eliminate more efficient future matches, while excessive delays raise the risk that orders expire before fulfillment. These trade-offs require coordinated decision-making that considers both immediate and long-term impact. As a result, a myopic approach that optimizes pricing or matching in isolation is likely to under-perform. An effective solution must jointly optimize pricing and assignment over time, adapting to the evolving system state and accounting for uncertainty in both order and shopper arrivals, as well as acceptance behavior.

\begin{figure}[!ht]
\begin{center}
\includegraphics[width=0.99\textwidth]{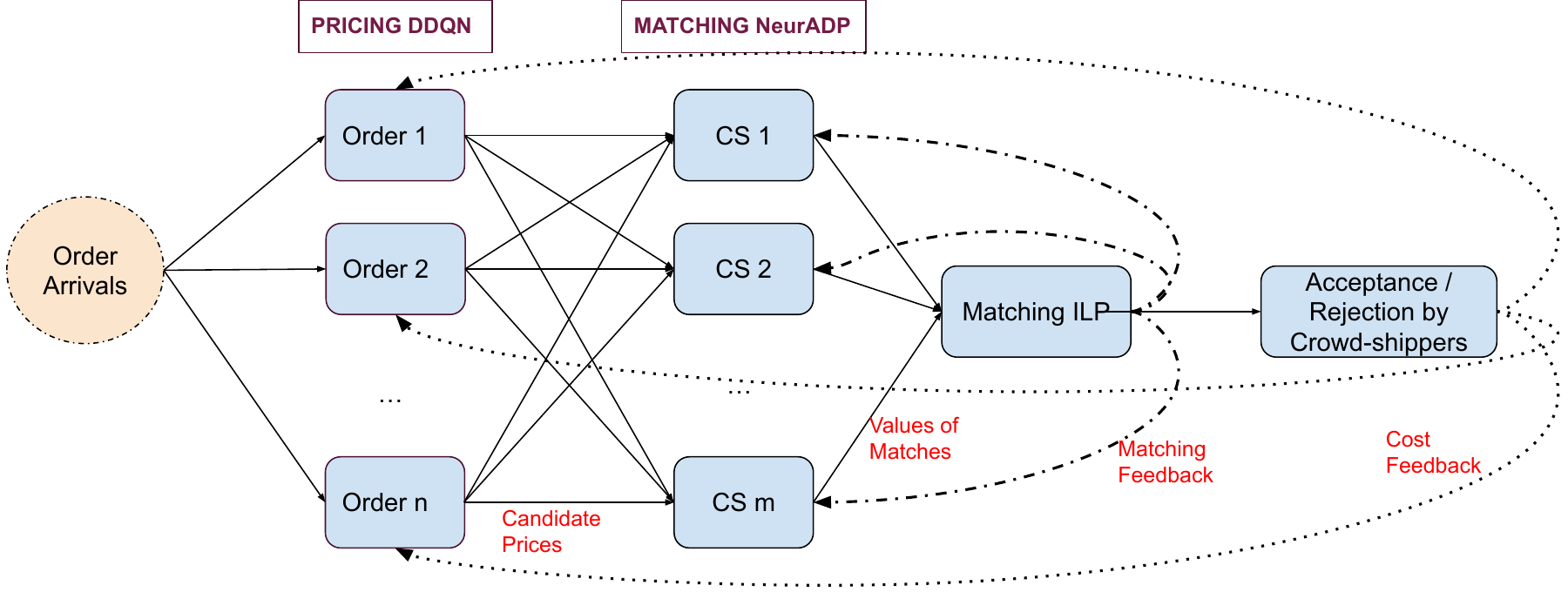}
\end{center}
\caption{Overview of the integrated pricing and matching framework, wherein `CS' stands for crowd-shipper.}
\label{fig:CSupdating_dynamics}
\end{figure}

We adopt an integrated framework that couples a DDQN model for pricing with a NeurADP-based matching algorithm. This joint approach allows pricing and assignment decisions to be updated in tandem, ensuring that the value function used in matching reflects the downstream cost implications of pricing decisions, while pricing learns to account for future matching feasibility and cost. By doing so, the system can adaptively balance short-term gains with long-term operational efficiency. Rather than optimizing each component in isolation, the two modules share feedback from crowd-shipper responses and matching outcomes, enabling coordinated learning that aligns pricing incentives with feasible and cost-effective assignments. Figure~\ref{fig:CSupdating_dynamics} illustrates this integrated flow: orders are priced by the DDQN network, candidate prices inform order–crowd-shipper pairing attractiveness, and match values are computed using the NeurADP value function. These values are used in a matching integer linear program (ILP), whose output is subject to acceptance by crowd-shippers. Feedback from both pricing and matching stages is looped back to update both learning modules, allowing the system to continually improve performance in a dynamic and uncertain environment.

In this section, we begin by introducing each component of our joint solution strategy. We first describe the DDQN pricing mechanism, detailing how it is integrated into the system and how its parameters are updated. We then present the NeurADP matching strategy, outlining its integration and learning dynamics. Finally, we bring these components together by describing the full NeurADP + DDQN framework, emphasizing how the two components interact and jointly contribute to the overall decision-making process.

% %%%%%%%%%%%%%%%%%
% %%%%%%%%%%%%%%%%%

\subsection{DDQN Pricing Solution}
\label{sec: CSpricing-ddqn-section}

To determine the pricing multiplier $\CSPricingIndividualMultiplier_{\CSIndividualOrderRequest}$ for each order $\CSIndividualOrderRequest \in \CSOrderRequestSet$, we incorporate a DDQN-based approach that estimates the expected downstream cost of different pricing decisions. The model evaluates the impact of each multiplier not only on immediate assignment outcomes but also on future system flexibility. Pricing decisions are guided by a learned action-value function \( Q(\CSIndividualOrderRequest, p) \), which estimates the long-run cost of assigning pricing multiplier \( p \) to order \( \CSIndividualOrderRequest \). The Q-function takes as input the state information of each order, which encodes its relevant attributes and contextual features at time $\CSCurrentTime$, and outputs a Q-value for each available pricing multiplier \( p \in \CSPricingDecisionSet \). The store then selects the multiplier that minimizes the estimated long-run cost:
\begin{equation}
\CSPricingIndividualMultiplier^\star_{\CSIndividualOrderRequest} = \min_{\CSPricingIndividualMultiplier \in \CSPricingDecisionSet} Q(\CSIndividualOrderRequest, \CSPricingIndividualMultiplier).
\end{equation}

The Q-function is trained through a decoupled evaluation and target estimation process. After each pricing decision and its observed outcome, the model updates its estimates by combining the immediate cost with the lowest predicted future cost, as computed by a separate target network. This target network is updated incrementally through a soft update rule that tracks the main Q-function. The overall setup supports stable learning and allows the pricing policy to capture both short-term effects and longer-term consequences. Further details regarding the pricing neural network and updating mechanism are provided in Section~\ref{sec:CStraining-details}.

\subsection{NeurADP Matching Solution}
\label{sec: CSmatching-neuradp-section}

NeurADP, introduced by \citet{shah2020neural}, is a value-based reinforcement learning framework for solving complex sequential decision problems. Its key innovation is embedding a neural network-based value function into an integer program that governs decision-making, allowing the system to incorporate long-term foresight without explicitly evaluating the vast number of potential future outcomes typical in dynamic environments. Unlike classical ADP, which relies on hand-crafted linear or piecewise-linear value approximations, NeurADP uses a neural network to automatically learn a compact, low-dimensional representation of the high-dimensional post-decision state space, removing the need for manual feature engineering. We utilize NeurADP rather than ADP, as proposed by \citet{mousavi2024approximate}, within our matching baselines due to the structural limitations inherent in their formulation. Their approach employs dual-value function approximation over aggregated spatial zones, which necessitates coarse spatial discretization and restricts crowd-shippers to serving only one zone that is not their home location. While this design choice aids in computational tractability, it imposes strong constraints on crowd-shipper behavior and reduces the expressiveness of the environment. In contrast, our NeurADP framework models interactions at the level of individual locations and allows crowd-shippers to serve multiple destinations, enabling a richer and more realistic representation of system dynamics. NeurADP is particularly well-suited to complex assignment problems, such as the many-to-one matching of orders to crowd-shippers considered in this paper. Its effectiveness in transportation applications has been demonstrated in prior work \citep{dehghan2023neural, shah2020neural, zhang2023future}. As such, we adopt a NeurADP-based approach to address the matching component of our crowd-shipping problem.

In our approach, we formulate the crowd-shipper to order matching problem as an ILP, which is solved at each time-step $\CSCurrentTime$. The ILP's objective integrates both immediate and future cost considerations, allowing decisions to account for downstream effects. To effectively capture the long-term impact of current matching and delay decisions, we augment the immediate cost evaluations with downstream value estimates that quantify the future cost implications of each action within the objective of our ILP. These future cost estimates are learned through a value function approximation trained using our NeurADP approach. The value function predicts the expected future cost from the next system state, allowing the model to make more forward-looking decisions at each decision point. We define the downstream value for each order as the estimated cost associated with its future trajectory, conditioned on its current state $\CSIndividualOrderRequest$. These estimates are computed using a neural network that takes the post-decision state of the order as input and returns a scalar value representing the expected future cost.

For a delay action on an order $\CSIndividualOrderRequest$, the downstream value is incorporated only if the order still has time remaining to be matched. Specifically, if $\CSIndividualOrderRequest_\CSRequestEpochsRemaining > 0$, the delay action incurs no immediate cost but carries forward a downstream value $\CSValueName(\CSIndividualOrderRequest)$, which reflects the potential future cost of needing to match or delay the order in later epochs. In contrast, if $\CSIndividualOrderRequest_\CSRequestEpochsRemaining = 0$, the order exits the system unserved, and the downstream value is zero. The updated cost of a delay action is therefore the sum of its immediate and downstream components:
\begin{equation}
\CSDelayCost_{\CSIndividualOrderRequest}^{\CSValueName} =
\CSDelayCost_{\CSIndividualOrderRequest} + \CSValueName(\CSIndividualOrderRequest),
\label{eq:CSdelay-total-cost}
\end{equation}
Subsequently, for a feasible matching $(\CSSingleBatch, \CSIndividualCrowdShipper)$, the downstream value depends on whether the offer is accepted or rejected. If accepted, the orders in the batch are fulfilled and exit the system, incurring no downstream cost. However, if the offer is rejected, the orders remain in the system and continue to incur costs in future epochs. We model this using a probabilistic expectation over the two outcomes:
\begin{equation}
\CSExpectedCost_{\CSSingleBatch \CSIndividualCrowdShipper}^{\CSValueName} = 
\CSAcceptProbFunc(\CSSingleBatch) \cdot \CSMatchCost_{\CSSingleBatch \CSIndividualCrowdShipper} + 
\left(1 - \CSAcceptProbFunc(\CSSingleBatch)\right) \cdot 
\left( \CSDelayCost_{\CSSingleBatch} + \sum_{\CSIndividualOrderRequest \in \CSSingleBatch} \CSValueName(\CSIndividualOrderRequest) \right),
\label{eq:CSexpected-cost-downstream}
\end{equation}
where the summation captures the aggregate downstream value of all orders in the batch under the rejection scenario. 

Thus, given the set of feasible matching and delay actions, we now formalize the store's decision-making process at each decision epoch $\CSCurrentTime \in \CSTime$ using an ILP formulation which captures how the store selects which orders to assign for delivery and which to delay, taking into account both expected delivery costs and potential delay penalties. We define two sets of binary decision variables. The variable $\CSActionVar_{\CSCurrentTime \CSSingleBatch \CSIndividualCrowdShipper} \in \{0,1\}$ indicates whether batch $\CSSingleBatch$ is assigned to crowd-shipper $\CSIndividualCrowdShipper$ at time $\CSCurrentTime$, where a value of 1 represents a selected assignment and 0 otherwise. Similarly, the variable $\CSDelayVar_{\CSCurrentTime \CSIndividualOrderRequest} \in \{0,1\}$ indicates whether order $\CSIndividualOrderRequest$ is delayed at time $\CSCurrentTime$. The ILP is formulated as follows:
\begin{subequations}
\begin{align}
\min_{\CSActionVar, \CSDelayVar} \quad & \sum_{(b,c) \in \CSFeasibleMatchingSet} \CSActionVar_{\CSCurrentTime \CSSingleBatch \CSIndividualCrowdShipper} \cdot \CSExpectedCost^{\CSValueName}_{\CSSingleBatch \CSIndividualCrowdShipper} 
+ \sum_{\CSIndividualOrderRequest \in \CSOrderRequestSet} \CSDelayVar_{\CSCurrentTime \CSIndividualOrderRequest} \cdot \CSDelayCost^{\CSValueName}_\CSIndividualOrderRequest \label{eq:CSilp-objective-downstream} \\
\text{s.t.} & \sum_{\substack{
    \CSIndividualCrowdShipper \in \CSCrowdShipperSet_\CSCurrentTime,
    \CSSingleBatch \in \CSPossibleBatchesSet_{\CSCurrentTime}: \\
    \CSIndividualOrderRequest \in \CSSingleBatch,
    (\CSSingleBatch, \CSIndividualCrowdShipper) \in \CSFeasibleMatchingSet_\CSCurrentTime
}} 
\CSActionVar_{\CSCurrentTime \CSSingleBatch \CSIndividualCrowdShipper} 
+ \CSDelayVar_{\CSCurrentTime \CSIndividualOrderRequest} = 1 
&& \hspace{-4.0em} \forall \CSIndividualOrderRequest \in \CSOrderRequestSet_\CSCurrentTime
\label{eq:CSorder_assignment} \\
& \sum_{\substack{
    \CSSingleBatch \in \CSPossibleBatchesSet_{\CSCurrentTime}: \\
    (\CSSingleBatch, \CSIndividualCrowdShipper) \in \CSFeasibleMatchingSet_\CSCurrentTime
}} 
\CSActionVar_{\CSCurrentTime \CSSingleBatch \CSIndividualCrowdShipper} \leq 1 
&& \hspace{-4.0em} \forall \CSIndividualCrowdShipper \in \CSCrowdShipperSet_\CSCurrentTime
\label{eq:CSshipper_capacity} \\
& \CSActionVar_{\CSCurrentTime \CSSingleBatch \CSIndividualCrowdShipper} \in \{0,1\} 
&& \hspace{-4.0em} \forall (\CSSingleBatch, \CSIndividualCrowdShipper) \in \CSFeasibleMatchingSet_\CSCurrentTime
\label{eq:CSaction_binary} \\
& \CSDelayVar_{\CSCurrentTime \CSIndividualOrderRequest} \in \{0,1\} 
&& \hspace{-4.0em}\forall \CSIndividualOrderRequest \in \CSOrderRequestSet_\CSCurrentTime
\label{eq:CSdelay_binary}
\end{align}
\end{subequations}
Constraint~\eqref{eq:CSorder_assignment} ensures that each order is either assigned to a feasible match or delayed, but not both. Constraint~\eqref{eq:CSshipper_capacity} ensures that each crowd-shipper is assigned at most one batch per decision epoch. Constraints~\eqref{eq:CSaction_binary} and~\eqref{eq:CSdelay_binary} enforce the binary nature of the decision variables. Finally, the objective~\eqref{eq:CSilp-objective-downstream} allows the ILP to reason over both short and long-term consequences of its matching decisions. By incorporating downstream value estimates, the decision-making process is guided toward solutions that reduce cumulative cost over time, rather than simply minimizing immediate penalties.

To update the value function approximations for our order-based post-decision states, we use automatic differentiation to compute gradients with respect to the value network parameters. These parameters are optimized by minimizing the loss between the estimated value of the post-decision states and the realized objective values, which are obtained from target values following ILP-based assignments. This allows the model to better estimate future rewards for each order based on actual downstream outcomes. To improve learning stability and computational efficiency, we apply off-policy updates, reusing experiences collected under previous policies without introducing instability into the value updates. We also introduce Gaussian noise during training to encourage exploration in a controlled manner. We provide further details of our matching neural network architecture and training process in Section~\ref{sec:CStraining-details}.

\subsection{NeurADP + DDQN Framework}
\label{sec:CSneuradp-ddqn-section}

We present our NeurADP + DDQN framework in Algorithm~\ref{alg:CSneuradp_ddqn} to jointly optimize pricing and matching decisions in our crowd-shipping system. At the start of each episode, the simulation environment is reset, and the set of retained orders is initialized. At every decision epoch, the system observes newly arriving orders, newly available crowd-shippers, and delayed orders carried forward from the previous epoch. These new and delayed orders are merged, and together with the crowd-shippers, they form the system state, denoted as $\CSStateNote_\CSCurrentTime$. For each outstanding order, we select a pricing multiplier using an $\epsilon$-greedy DDQN policy, which relies on Q-value predictions generated by a neural network with parameters $\theta^{Q}$. After determining these pricing decisions, we identify all feasible order–shopper pairings. We then solve the resulting assignment problem using Equation~\eqref{eq:CSilp-objective-downstream}. Downstream value estimates embedded within the ILP are computed via a value network parameterized by $\theta^{V}$. This network approximates the expected future costs associated with each order's post-decision state. Once assignments are made, shoppers decide whether to accept or reject their assigned orders. The store then incurs the associated delivery or delay costs. Orders not matched or deemed infeasible are either postponed or removed, prompting a state transition. During the training process, we maintain two separate experience replay buffers: one for pricing and another for matching. Experiences from each epoch are stored in the respective buffers, which periodically provide mini-batches used to update network parameters via stochastic gradient descent.

\begin{algorithm}[H]
\caption{NeurADP + DDQN Framework}
\label{alg:CSneuradp_ddqn}
\begin{algorithmic}[1]
\State Initialize replay buffers $\mathcal{M}^{\texttt{Pricing}}$ and $\mathcal{M}^{\texttt{Matching}}$, and network parameters $\theta^{Q}$ and $\theta^{V}$

\For{episode $\CSEpisode = 1,\ldots,\CSEpisodeSet$}
  \State Reset store environment, $\CSOrderRetained_0=\emptyset$
  \For{decision epoch $\CSCurrentTime = 1,\ldots,T$}
    \State Observe arrivals $\CSExogenousInformation_{\CSCurrentTime}^{\CSOrd}$, $\CSExogenousInformation_{\CSCurrentTime}^{\CSCrowd}$, and delayed orders $\CSOrderRetained_{\CSCurrentTime-1}$
    \State Form state $\CSStateNote_{\CSCurrentTime} = \bigl(\CSCrowdShipperState_{\CSCurrentTime},\CSRequestState_{\CSCurrentTime}\bigr)$
    \State Select multiplier $\CSPricingIndividualMultiplier_{\CSIndividualOrderRequest}$ using $\epsilon$–greedy DDQN with parameters $\theta^{Q}$ for each $\CSIndividualOrderRequest \in \CSOrderRequestSet_{\CSCurrentTime}$
    \State Build feasible matching set $\CSFeasibleMatchingSet_{\CSCurrentTime}$
    \If{training pricing}
        \State Store pricing experience $e^\texttt{Pricing}$ in $\mathcal{M}^{\texttt{Pricing}}$
        \State Sample mini-batches from $\mathcal{M}^{\texttt{Pricing}}$, update DDQN parameters $\theta^{Q}$
    \EndIf
    \State Solve ILP~\eqref{eq:CSilp-objective-downstream} and add Gaussian noise for exploration
    \If{training matching}
        \State Store matching experience $e^\texttt{Matching}$ in $\mathcal{M}^{\texttt{Matching}}$
        \State Sample mini-batches from $\mathcal{M}^{\texttt{Matching}}$, update value network parameters $\theta^{V}$
    \EndIf
    \State Realize endogenous crowd-shipper acceptance $\CSExogenousInformation^\CSAcc$
    \State Execute actions, incur cost $\CSCostName_{\CSCurrentTime}$, delay orders $\CSOrderRetained_\CSCurrentTime$ 
    \State Transition to subsequent state $\CSStateNote_{\CSCurrentTime+1}$
  \EndFor
\EndFor
\end{algorithmic}
\end{algorithm}

%%%%%%%%%%%%

%%%%%%%%%%%%%%%%%%%%%%%%%%%%%%%%%%%%%%%%%
%%%% Subsection: Dataset Description %%%%
%%%%%%%%%%%%%%%%%%%%%%%%%%%%%%%%%%%%%%%%%
\section{Experimental Setup} \label{CSexperimentalsetup}
We implement our methods in Python 3.10.9 and conduct numerical experiments on Google Cloud virtual machines, utilizing both an e2-standard-8 instance (8 vCPUs, 32 GB memory) and an e2-highmem-16 instance (16 vCPUs, 128 GB memory), each based on x86/64 architecture and AMD Rome CPUs. All ILP models are solved using IBM ILOG CPLEX Optimization Studio, version 22.1.1. Below, we present a detailed overview of the experimental settings used to evaluate the effectiveness of our matching and pricing policies. This includes a comprehensive description of the temporal and geographic distributions of incoming orders and crowd-shippers, the architectures and training processes of the neural networks used for matching and pricing, and the benchmark policies employed in our comparative analysis.

\subsection{Dataset Description}
\begin{figure}[!ht]
\begin{center}
\includegraphics[width=0.99\textwidth]{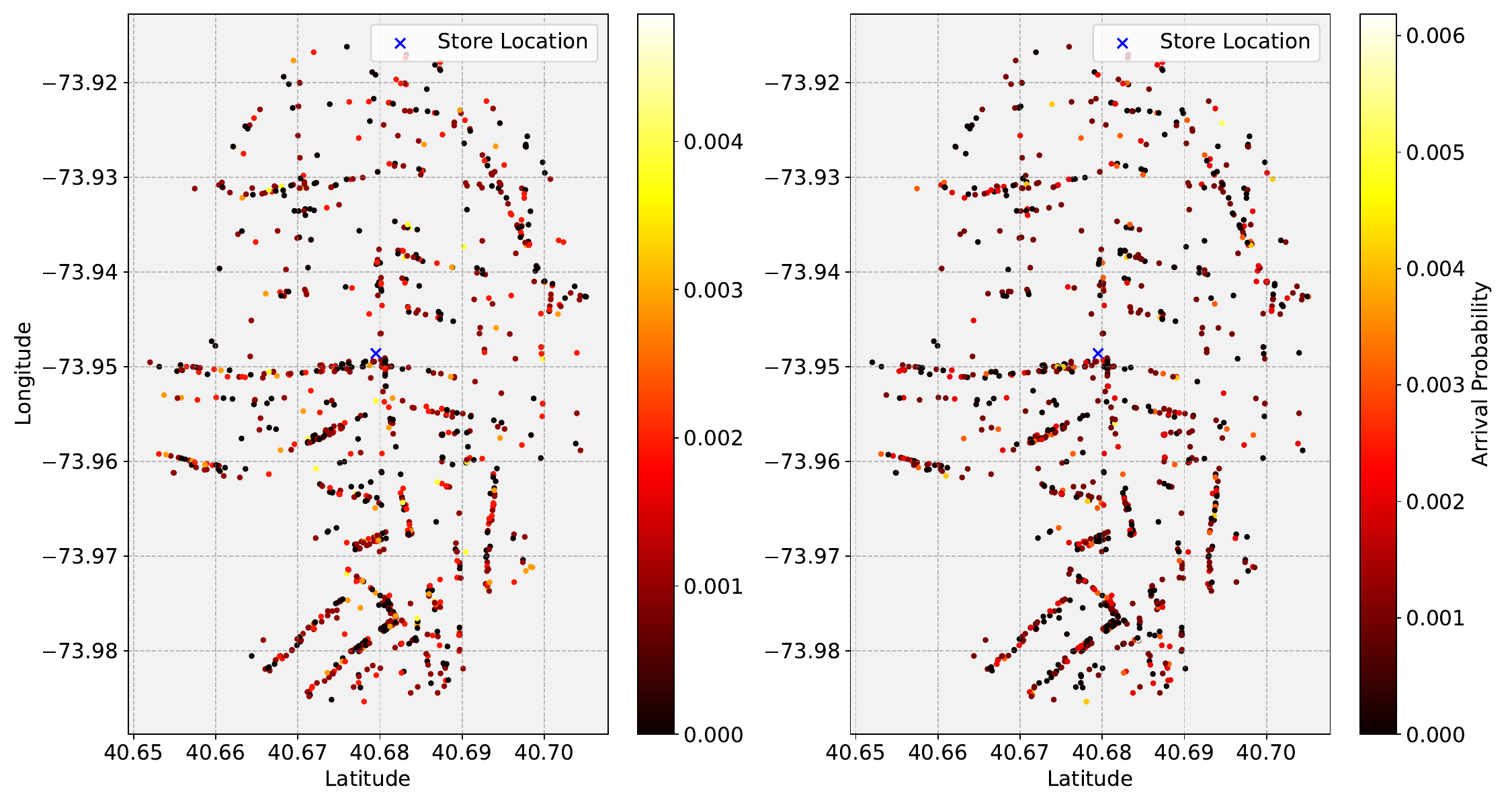}
\end{center}
\caption{Geographic arrival distribution of order destinations (left) and crowd-shipper home locations (right) at $t=0$ decision epoch.}
\label{fig:CSOrderDistribution_Geographic}
\end{figure}

In our numerical study, we utilize a real-world dataset representing online delivery activity in Brooklyn, USA. The dataset includes latitude-longitude coordinates for 988 unique drop-off locations, each corresponding to a historical delivery request. These coordinates are treated as potential delivery destinations and home locations and are used to construct time-varying, location-based probability distributions for both orders and crowd-shippers. For each decision epoch, these spatial distributions are generated independently using Poisson sampling to capture spatial and temporal variability in demand and supply. Although both distributions are derived using the same generative method, they are sampled separately, resulting in distinct spatial patterns for orders and crowd-shippers. This reflects real-world dynamics, where customer availability (e.g., crowd-shippers) and order demand often originate from different parts of a city such as residential neighborhoods versus commercial zones. A visual example of these spatial distributions at the start of the planning horizon is shown in Figure~\ref{fig:CSOrderDistribution_Geographic}.

\begin{figure}[!ht]
\begin{center}
\includegraphics[width=0.99\textwidth]{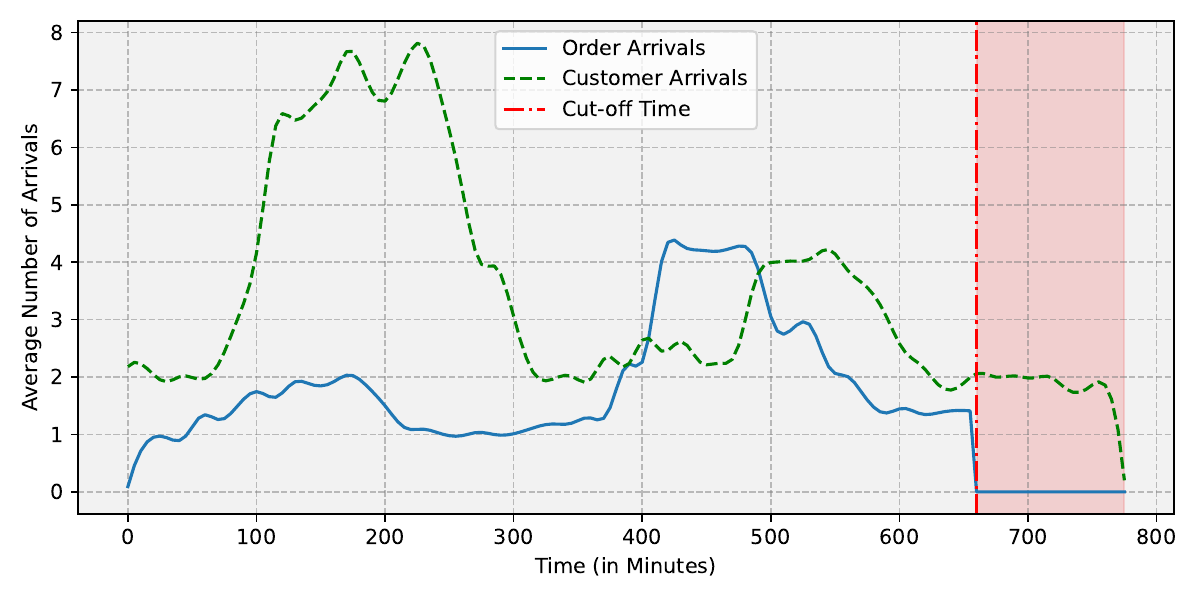}
\end{center}
\caption{Order and crowd-shipper arrival distribution throughout work day.}
\label{fig:CSOrderDistribution_Arrival}
\end{figure}

A single store is positioned at the center of the delivery region and serves as the origin for all deliveries. The planning horizon spans 13 hours and is divided into 5-minute decision epochs (i.e., $\CSTimeIntervals = 5$ minutes), resulting in 156 decision points throughout the day. At each decision epoch, the number of incoming orders and customers is determined using smoothed and interpolated arrival profiles derived from real-world data. These profiles are constructed from predefined time series that reflect realistic hourly fluctuations in demand and crowd-shipper availability. To introduce controlled stochasticity, the number of arrivals at each epoch is sampled from a normal distribution centered around the interpolated mean arrival rate, i.e., $\mathcal{N}(\mu, \sigma^2)$, where $\mu$ is the expected number of arrivals and $\sigma$ is set to 1 by default. The sampled values are then rounded and clipped at zero to ensure non-negativity. A cut-off is imposed toward the end of the planning horizon with 2 hours remaining, after which no new orders are introduced. We provide the temporal distribution of order and crowd-shipper arrivals in Figure~\ref{fig:CSOrderDistribution_Arrival}.

% %%%%%%%%%%%%%%%%%
% %%%%%%%%%%%%%%%%%

% %%%%%%%%%%%%%%%%%
% %%%%%%%%%%%%%%%%%

%%%%%%%%%%%%%%%%%%%%%%%%%%%%%%%%%%%%%%%%%
%%%% Subsection:  Neural Network Architecture and Training %%%%
%%%%%%%%%%%%%%%%%%%%%%%%%%%%%%%%%%%%%%%%%
\subsection{Neural Network Architecture and Training}
\label{sec:CStraining-details}

Our framework employs two separate feedforward neural networks to support value estimation for matching and pricing decisions, respectively. Both models share a common architectural backbone composed of fully connected layers with Exponential Linear Unit (ELU) activations, differing primarily in input structure and decision context.

For the matching task, the network predicts the desirability of assigning an order under a given system state. The input consists of four features: (i) a categorical indicator of the destination location, (ii) the normalized number of decision epochs remaining until the deadline, and (iii) the normalized current time in the planning horizon. The destination is first passed through a learnable embedding layer to capture latent spatial correlations. This embedding is concatenated with the three continuous features and passed through three fully connected layers, each with 300 hidden units and ELU activations. The output layer is a single linear unit that returns a scalar value representing the estimated utility of matching the order in the current state. This design enables the network to flexibly model interactions between spatial, temporal, and deadline-related factors. The pricing network extends this architecture by additionally conditioning on a candidate pricing action. Each input vector includes: (i) the destination location of the order, (ii) the normalized number of epochs remaining until the deadline, (iii) the normalized current time, (iv) the number of candidate orders currently being considered, and (v) a one-hot encoding of a pricing multiplier selected from the discrete set $\{0.8, 0.9, 1.0, 1.1, 1.2\}$. The destination input is embedded and concatenated with the remaining inputs before being passed through the same feedforward architecture used in the matching model. The final scalar output represents the estimated cost associated with assigning the corresponding pricing multiplier.

These neural networks are trained concurrently within a reinforcement learning framework using experience replay and soft target updates. During each decision epoch, the system interacts with a simulated environment to generate observations, compute action values, and execute decisions. These interactions are stored in prioritized replay buffers and periodically sampled to update the network parameters. For the matching network, each training experience captures the state of the system, a set of feasible assignments, and the associated cost signals. The network is trained to minimize the gap between its predicted cost-to-go values and targets constructed from both immediate costs and the expected future costs. A temporal-difference learning approach is used, with target values computed via a separate target network to enhance training stability. Gradients are propagated using a mean-squared error loss, and a prioritization mechanism ensures that high-value or high-error experiences are sampled more frequently. The pricing network is trained analogously. Each experience consists of a current order state, a set of pricing actions, and their respective outcomes in terms of customer acceptance and incurred cost. The model learns to score each pricing level based on its expected cost using temporal-difference targets. To encourage exploration, pricing actions are stochastically sampled during training via a softmax distribution weighted by predicted cost. As in the matching case, the pricing model employs soft target network updates and an adaptive learning rate schedule for improved convergence behavior. Together, these training mechanisms allow both neural networks to learn temporally extended value functions under operational uncertainty, enabling flexible and high-quality decision-making throughout the planning horizon.

%%%%%%%%%%%%%%%%%%%%%%%%%%%%%%%%%%%%%%%%%
%%%% Subsection:  Baseline Policies %%%%
%%%%%%%%%%%%%%%%%%%%%%%%%%%%%%%%%%%%%%%%%
\subsection{Baseline Policies}
We consider three baseline policies to compare against our \CSND~approach. 
\begin{itemize}
    \item \CSNF: This method employs the NeurADP framework to learn the matching policy while using a fixed pricing strategy. To determine a robust fixed pricing level, we evaluate five candidate multipliers from the set $\{0.8, 0.9, 1.0, 1.1, 1.2\}$, each paired with the trained matching network. For each multiplier, we simulate a full operational day using identical order and customer arrivals, and compute the total incurred cost. The fixed multiplier that yields the lowest cost across simulations is then selected. Empirically, we find that the $1.0$ multiplier consistently performs best given the arrival distributions and cost structure, and thus adopt it as the default for the \CSNF~policy in all forthcoming evaluations. We note that this baseline can be considered as an enhanced version of \citet{mousavi2024approximate}'s approach.

    \item \CSGD: In this approach, a greedy matching procedure is utilized alongside a DDQN-based pricing policy. At each decision epoch, the greedy matching algorithm prioritizes orders based on urgency, favoring those with fewer remaining epochs, and sequentially assigns them to available customers. A feasible match must involve an unassigned customer and unassigned orders, and is evaluated based on its associated detour cost. The action with the lowest detour cost is selected; if no viable match exists, the order is delayed. All matched entities are marked unavailable for further assignments within the same epoch. While heuristic, this approach balances tractability with responsiveness and serves as a practical benchmark for real-time decision-making.

    \item \CSGF: This approach corresponds to the most naive baseline, combining the same greedy matching heuristic with the fixed pricing multiplier of $1.0$. 
\end{itemize}
Together, these baselines help isolate the contributions of NeurADP and DDQN, both individually and jointly, and serve as meaningful comparators in assessing the value of more sophisticated decision policies.

% We utilize NeurADP rather than ADP, as proposed by \citet{mousavi2024approximate}, within our matching baselines due to the structural limitations inherent in their formulation. Their approach employs dual-value function approximation over aggregated spatial zones, which necessitates coarse spatial discretization and restricts crowd-shippers to serving only one zone that is not their home location. While this design choice aids in computational tractability, it imposes strong constraints on crowd-shipper behavior and reduces the expressiveness of the environment. In contrast, our NeurADP framework models interactions at the level of individual locations and allows agents to serve multiple destinations, enabling a richer and more realistic representation of system dynamics.

%%%%%%%%%%%%%%%%%%%%%%%%%%%%%%%%%%%%%%%%%
%%%% Subsection: Results %%%%
%%%%%%%%%%%%%%%%%%%%%%%%%%%%%%%%%%%%%%%%%
\section{Results} \label{CSresults}

In this section, we present the results of our detailed numerical study for the crowd-shipping problem. Our primary focus is on evaluating the average cost incurred over a 13-hour decision horizon, comparing the performance of the \CSND~policy against three benchmark policy classes. Each policy is evaluated using 50 days of testing data, with the entire 50-day set repeated 5 times to compute reliable averages and standard deviations. At each decision epoch, new orders are sampled and combined with any previously delayed orders. Based on the current policy, a feasible batching and matching decision is made, pairing orders with available crowd-shippers. Once matched, the crowd-shippers immediately leave the store, deliver the assigned orders to their destinations, and then return home. The analysis below explores the overall performance of the proposed and benchmark policies, highlighting key trends and managerial insights.

\subsection{Crowd-shipping Performance}

We evaluate the performance of our policies across five key input parameters: the allowed delay time, the cost of deviation from a crowd-shipper's direct route home, the base compensation per delivery, the number of unique delivery locations permitted, and the ratio of crowd-shippers to orders. The delay time refers to the maximum amount of time an order can remain in the system before being fulfilled by traditional delivery methods if unmatched. The deviation cost represents the per-minute compensation paid to crowd-shippers for deviating from their direct route home. For example, if a crowd-shipper's normal route home takes 10 minutes and their delivery route takes 15, they are compensated for the 5-minute deviation, with the per-minute rate varied across experiments. The base compensation is the fixed amount paid to a crowd-shipper for each order delivered, regardless of delay or deviation. The number of unique delivery locations defines how many distinct addresses a crowd-shipper may deliver to before returning home, e.g., a value of 2 means a shipper can serve two separate destinations per trip. Lastly, the crowd-shipper-to-order ratio indicates the relative supply of crowd-shippers to demand from incoming orders over a given day. In our default configuration, we allow a 90-minute delay window, limit crowd-shippers to one delivery location, and set the crowd-shipper-to-order ratio to approximately 0.45. We impose an \$8 penalty for un-served orders, offer a base compensation of \$4, and use a deviation multiplier of 0.1 (i.e., 10 cents per minute of route deviation). In what follows, we explore the performance of our policies against the aforementioned parameter variations. We provide further information regarding the run-time and training in Section~\ref{sec:CSalgperformance}.

\subsubsection{Variation of Deviation Cost Multipliers}

We begin by evaluating the impact of the deviation payment rate, which determines how much additional compensation crowd-shippers receive for each minute they deviate from their home locations when making deliveries. Table~\ref{table:CSdeviation_payment_comparison} summarizes the cost performance of each policy across deviation payment rates of \{0.0, 0.1, 0.2, 0.3\}. The column labeled \CSND~reports the average total cost (with standard deviation) incurred under our \CSND~policy, computed over five iterations of 50 test days. The columns labeled \textit{\% Over (N+F)}, \textit{\% Over (G+D)}, and \textit{\% Over (G+F)} denote the relative percentage cost saved for \CSND~compared to each baseline policy. These are calculated as $(\text{Baseline} - \CSND) / \CSND$, and express how much more expensive each baseline is relative to our proposed \CSND~approach. These abbreviations and computation conventions are used consistently throughout the remainder of the empirical analysis. 

As expected, total cost increases with higher deviation payments, as crowd-shippers are compensated more for each additional minute of deviation. Across all settings, \CSND~consistently achieves the lowest cost, while \CSGF~incurs the highest. \CSGD~performs better than \CSGF, but remains worse than both \CSND~and \CSNF. Notably, the cost advantage of \CSND~over greedy matching policies grows substantially as the deviation multiplier increases, from just 1.70\% and 2.63\% over \CSGD~and \CSGF~at a multiplier of 0.0, to 18.40\% and 18.71\% at 0.3, respectively. This reflects the increasing importance of smart matching as deviation costs become more prominent. Unlike greedy policies that maximize immediate matches without regard to future opportunities, \CSND~adopts an approach that strategically delays matches when it anticipates more efficient downstream options. This behavior allows \CSND~to minimize detours and reduce deviation-related costs more effectively as those costs rise.

\begin{figure}[!ht]
\begin{center}
\includegraphics[width=0.99\textwidth]{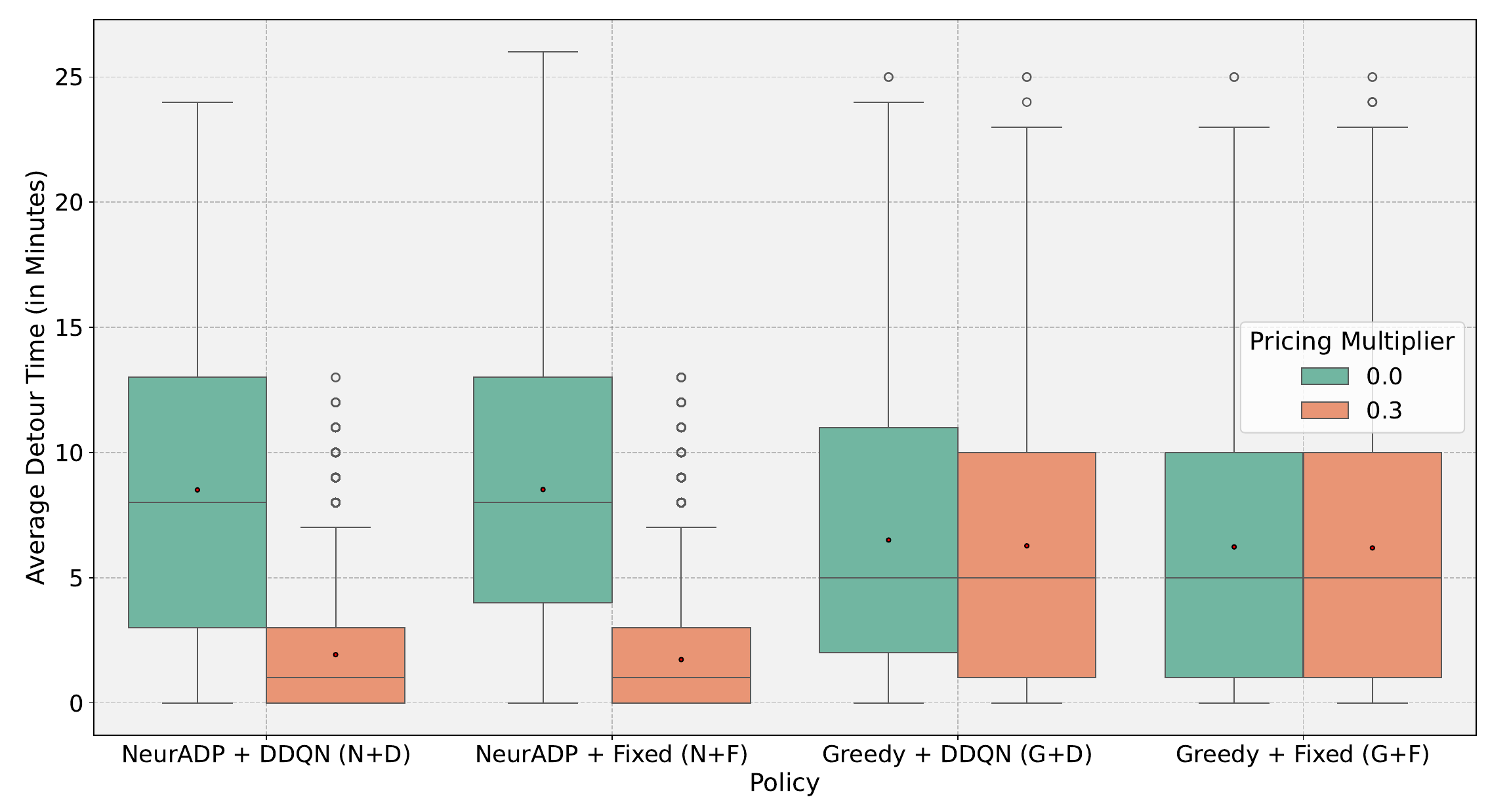}
\end{center}
\caption{Average deviation time (in minutes) for 0.0 and 0.3 deviation multiplier scenario for each policy.}
\label{fig:CSdeviation_mulltiplier_0_3}
\end{figure}

Interestingly, the cost gap between \CSND~and \CSNF~narrows as the deviation multiplier increases, dropping from 6.68\% at 0.0 to just 0.87\% at 0.3. This trend reveals a shift in what drives cost efficiency: when deviation payments are zero, pricing is the only form of compensation, making it a key lever for managing cost. In this setting, smart pricing strategies in \CSND~deliver notable benefits over fixed-pricing baselines. However, as the deviation multiplier increases, matching becomes the dominant driver of cost, and the marginal benefit of pricing diminishes. This behavioral shift is further evident in Figure~\ref{fig:CSdeviation_mulltiplier_0_3}, which shows the detour distance distributions across deviation multiplier settings. At a multiplier of 0.0, both \CSND~and \CSNF~incur high detour distances, as there is no penalty for deviating from a crowd-shipper's path, resulting in broader distributions and more outliers. In contrast, at a multiplier of 0.3, these policies become highly sensitive to deviation costs and adjust their matching behavior accordingly, prioritizing low-deviation matches and avoiding inefficient detours. The mean detour distance for \CSND~drops from 8.51 to 1.93 minutes, and for \CSNF~from 8.52 to 1.73, with corresponding reductions in median, interquartile range, and outliers. Meanwhile, the greedy policies remain largely unchanged: \CSGD~shifts only marginally from 6.50 to 6.27, and \CSGF~from 6.23 to 6.19. This lack of responsiveness highlights their inability to internalize new cost structures.

% %%%%%%%%%%%%%%%%%
% %%%%%%%%%%%%%%%%%

\begin{table}[!ht]
\centering
\caption{Average cost of \CSND~under varying deviation payments and its relative cost increase compared to benchmark policies.}
\label{table:CSdeviation_payment_comparison}
\small
\begin{tabular}{lcccc}
\toprule
\textbf{Deviation Payment} & \textbf{(N+D)} & \textbf{\% Over (N+F)} & \textbf{\% Over (G+D)} & \textbf{\% Over (G+F)} \\
\midrule
0.0 & \$1,156.43 $\pm$ 1.39 & +6.68\% & +1.70\% & +2.63\% \\
0.1 & \$1,256.78 $\pm$ 3.10 & +3.01\% & +3.05\% & +4.62\% \\
0.2 & \$1,297.94 $\pm$ 3.50 & +2.03\% & +10.74\% & +11.75\% \\
0.3 & \$1,331.24 $\pm$ 1.45 & +0.87\% & +18.40\% & +18.71\% \\
\bottomrule
\end{tabular}
\end{table}

\subsubsection{Variation of Base Compensation}

In the following, we assess the impact of base compensation provided to crowd-shippers on the overall cost incurred throughout the workday. We consider a range of base compensation levels, specifically \{\$2, \$3, \$4, \$5\}, with corresponding results presented in Table~\ref{table:CSbase_compensation_comparison}. As expected, total costs increase with higher compensation across all policies. However, \CSND~consistently achieves the lowest total cost at each compensation level, while the highest costs are generally incurred by the greedy matching policies \CSGD~and \CSGF. The relative savings provided by \CSND~over the other policies diminish as base compensation rises, dropping from double-digit percentages at \$2 to single digits at higher compensation levels. This trend suggests that \CSND's benefits from smarter matching and pricing decisions are most impactful when compensation is low. As base costs rise, the opportunity for strategic decisions to meaningfully reduce total cost narrows since every match carries a higher fixed cost, pricing to limit detours and optimize routing has a reduced effect on marginal savings.

\begin{figure}[!ht]
\begin{center}
\includegraphics[width=0.99\textwidth]{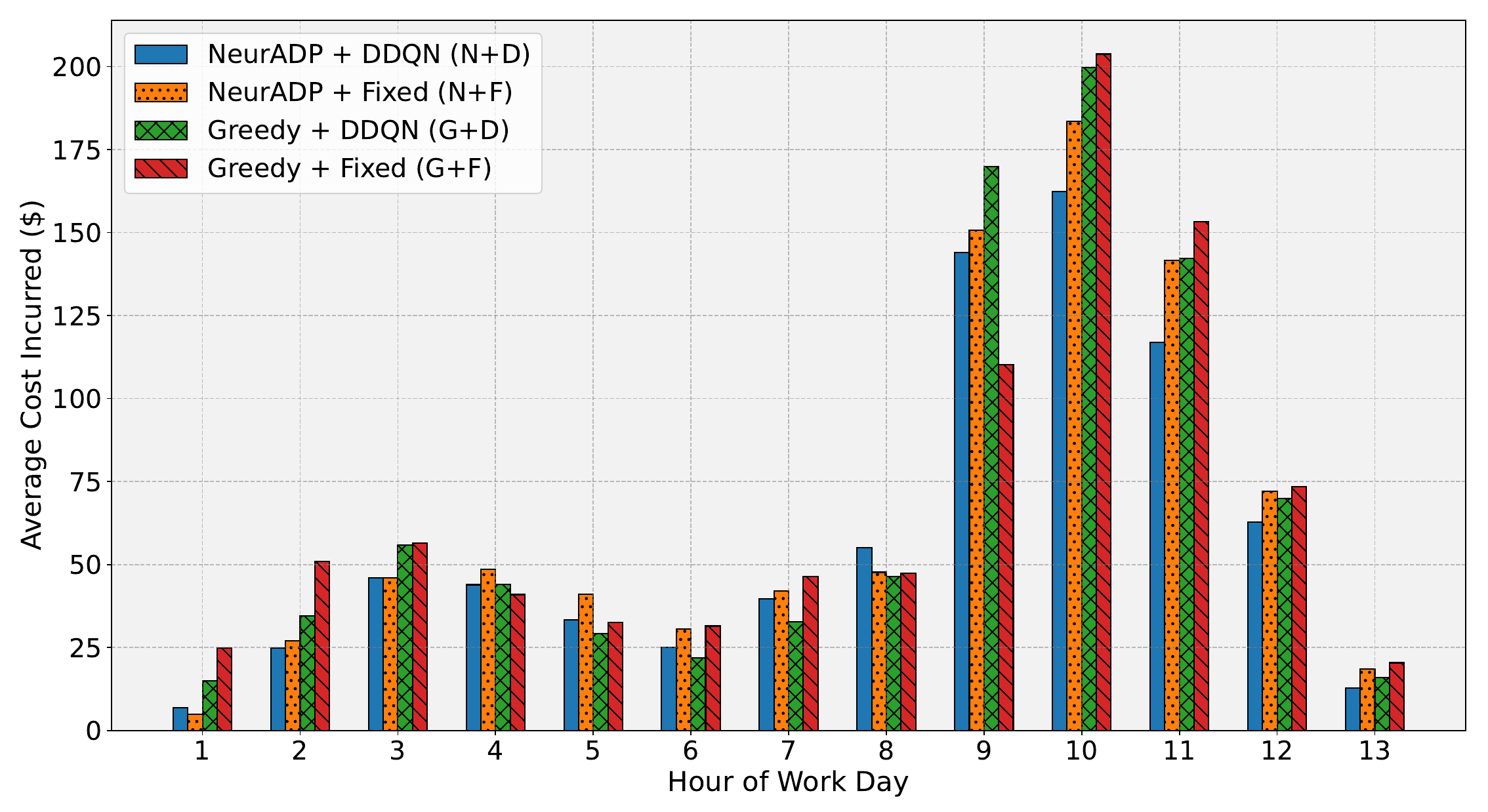}
\end{center}
\caption{Average cost incurred at each hour of the workday for each policy.}
\label{fig:CSbase_comp_day}
\end{figure}

Figure~\ref{fig:CSbase_comp_day} further illustrates this dynamic. Under the \$2 base compensation scenario, the majority of \CSND's advantage is concentrated during the later hours of the workday, particularly between hours 9 and 12, when order volume and delivery activity peak. During these high-demand periods, overall costs rise sharply across all policies; however, \CSND~demonstrates an ability to dampen these spikes significantly better than its counterparts. This effect is especially pronounced when compared to the \CSGD~and \CSGF~policies, which incur steep cost surges. The heightened savings during these rush hours reflect how intelligent decision-making under \CSND~can absorb the stress of high volume more effectively, avoiding inefficient assignments and mitigating cost escalation, thereby contributing disproportionately to the overall savings observed.

% %%%%%%%%%%%%%%%%%
% %%%%%%%%%%%%%%%%%

\begin{table}[!ht]
\centering
\caption{Average cost of \CSND~under varying base compensation values and its relative cost increase compared to benchmark policies.}
\label{table:CSbase_compensation_comparison}
\small
\begin{tabular}{lcccc}
\toprule
\textbf{Base Compensation} & \textbf{(N+D)} & \textbf{\% Over (N+F)} & \textbf{\% Over (G+D)} & \textbf{\% Over (G+F)} \\
\midrule
\$2.0 & \$776.51 $\pm$ 1.97 & +10.45\% & +13.18\% & +14.64\% \\
\$3.0 & \$1,030.26 $\pm$ 5.09 & +5.62\% & +6.31\% & +6.83\% \\
\$4.0 & \$1,256.78 $\pm$ 3.10 & +3.01\% & +3.05\% & +4.62\% \\
\$5.0 & \$1,437.25 $\pm$ 2.67 & +5.00\% & +3.70\% & +6.47\% \\
\bottomrule
\end{tabular}
\end{table}

\subsubsection{Variation of Unique Delivery Locations}

\begin{figure}[!ht]
\begin{center}
\includegraphics[width=0.99\textwidth]{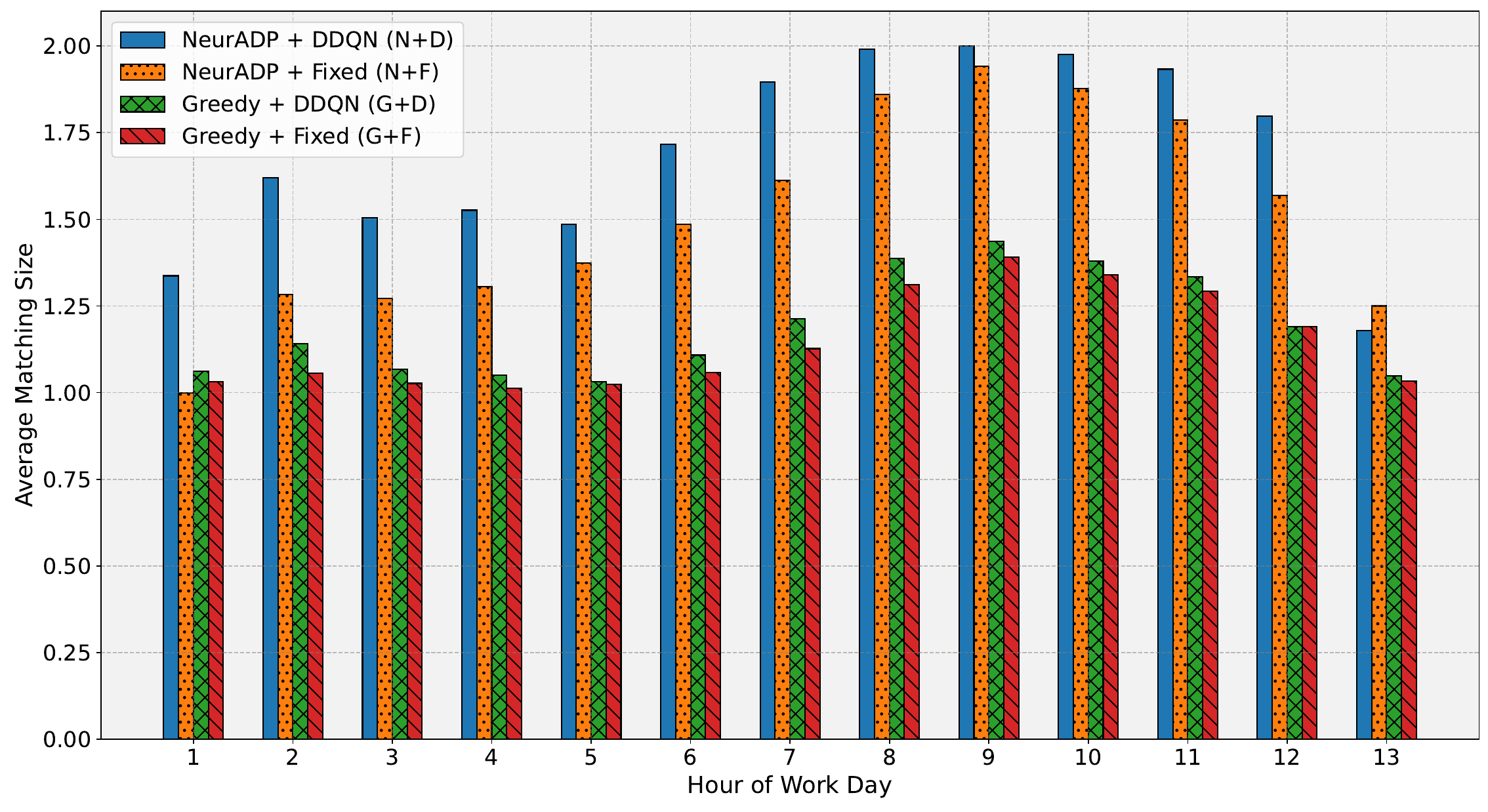}
\end{center}
\caption{Average number of orders matched to each crowd-shipper in the 2 unique locations scenario for each policy.}
\label{fig:CSmatching_size_over_day}
\end{figure}

We now examine how allowing additional detour locations influences cost and matching behavior. Table~\ref{table:CSunique_locations_comparison} shows that enabling two detour locations instead of one results in substantial cost reductions across all policies, with \CSND~achieving the largest absolute and relative gains. Specifically, the total cost incurred by \CSND~drops from \$1256.78 to \$1036.49 which is a reduction of over 17\%. This improvement leads to significantly higher savings over the other policies: \CSND~outperforms \CSNF~by 9.91\%, \CSGD~by 13.45\%, and \CSGF~by 14.98\%. This performance gap is largely driven by \CSND's superior ability to exploit the increased flexibility in matching. As shown in Figure~\ref{fig:CSmatching_size_over_day}, \CSND~achieves a higher average matching size throughout the day when two detour locations are allowed, reaching nearly 2.0 during peak hours, compared to only around 1.2 for \CSGD~and \CSGF. This indicates that \CSND~more effectively consolidates orders, leading to fewer trips and greater cost efficiency.

Supporting this, we see from the breakdown of cost components that \CSND~achieves the lowest total cost per time unit (6.66), compared to 7.32 for \CSNF, 7.53 for \CSGD, and 7.66 for \CSGF. The majority of \CSND's advantage stems from reduced matching costs (5.85 vs. 6.64 for \CSNF and 7.07 for \CSGF), as it is able to construct larger and more efficient matchings. Interestingly, although \CSND~incurs a slightly higher lost order cost than \CSNF~(0.81 vs. 0.69), this is more than offset by its savings in matching costs. Greedy policies, which do not plan across time, fail to leverage the added matching flexibility and show relatively small improvements when increasing detour locations, from a matching size of 1.0 to only around 1.2, and continue to perform poorly in terms of both total cost and cost breakdown.

% %%%%%%%%%%%%%%%%%
% %%%%%%%%%%%%%%%%%

\begin{table}[!ht]
\centering
\caption{Average cost of \CSND~for different numbers of unique delivery locations, and relative savings over benchmark policies.}
\label{table:CSunique_locations_comparison}
\small
\begin{tabular}{lcccc}
\toprule
\textbf{Unique Locations} & \textbf{(N+D)} & \textbf{\% Over (N+F)} & \textbf{\% Over (G+D)} & \textbf{\% Over (G+F)} \\
\midrule
1 & \$1,256.78 $\pm$ 3.10 & +3.01\% & +3.05\% & +4.62\% \\
2 & \$1,036.49 $\pm$ 6.58 & +9.91\% & +13.45\% & +14.98\% \\
\bottomrule
\end{tabular}
\end{table}

\subsubsection{Variation of Order to Crowd-shipper Ratio}

We now analyze how the \CSND~policy adapts its pricing and matching behavior under varying levels of supply-demand imbalance, using the order-to-crowd-shipper ratio as a proxy. Table~\ref{table:CSorder_customer_ratio_comparison} shows the total costs incurred by each policy at ratios of approximately 0.33, 0.45, and 0.60. Across all settings, \CSND~consistently achieves the lowest cost, with savings of up to 6.57\% compared to the worst-performing baseline (\CSGF). As the order-customer ratio increases from 0.33 to 0.60, the total cost for \CSND~rises from \$1011.31 to \$1470.96, reflecting the expected increase in system burden as more orders compete for fewer available crowd-shippers. However, the strength of \CSND~lies in its ability to adapt to this shift through dynamic pricing and strategic matching.

\begin{figure}[!ht]
\begin{center}
\includegraphics[width=0.99\textwidth]{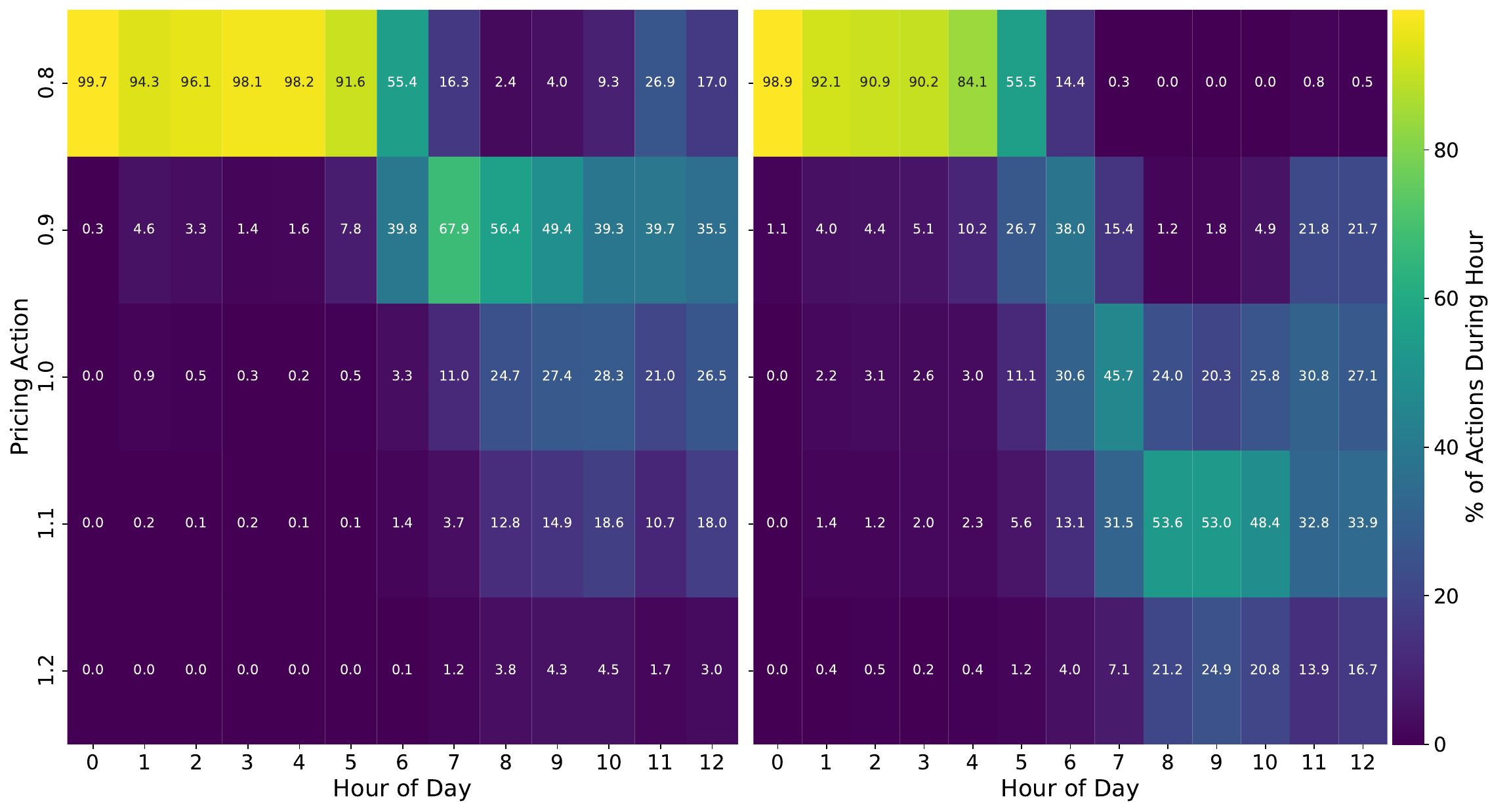}
\end{center}
\caption{Percentage of time each pricing multiplier action is taken in the 0.33 case (left) and 0.6 (right) scenario of order-crowdshipper ratio for \CSND~policy.}
\label{fig:CSpricing_action_comparison_heatmap}
\end{figure}

This adaptive behavior is visualized in Figure~\ref{fig:CSpricing_action_comparison_heatmap}, which compares the hourly pricing action distributions of \CSND~under the two extremes: 0.33 (left) and 0.60 (right). At a low order-customer ratio (0.33), the platform is crowd-shipper-rich, and \CSND~takes advantage of this by aggressively minimizing payouts, setting deviation pricing multipliers of 0.8 for nearly all actions in the early hours, reaching up to 99\% usage. Because future crowd-shipper availability is ample, \CSND~can afford to offer lower prices for deviation, knowing that the probability of failing to serve an order later in the day is minimal. It capitalizes on this flexibility to limit cost without sacrificing feasibility. As demand builds throughout the day, \CSND~gradually transitions to more moderate multipliers (0.9–1.0), but still avoids costly compensation.

In contrast, at a higher order-customer ratio of 0.60, where the platform faces a tighter supply of crowd-shippers, \CSND~responds with a markedly different pricing strategy. The use of 0.8 pricing drops precipitously, and mid-to-high pricing actions (1.0–1.2) become increasingly common, particularly during peak hours. This shift reflects the model's awareness of tightening feasibility constraints; it begins offering more generous payouts to incentivize matches and reduce the risk of order loss. Notably, pricing action 1.0 alone accounts for over 50\% of all actions in certain periods, with significant upticks in 1.1 and 1.2 as well. This behavioral pivot demonstrates how \CSND~prioritizes service reliability and match inducement when the delivery network becomes constrained.

Overall, this comparison highlights the core advantage of the \CSND~framework: its ability to learn and execute context-aware pricing strategies that are sensitive to real-time supply-demand conditions. Unlike static or greedy approaches, \CSND~uses learned policy signals to balance cost efficiency with service feasibility, adjusting its strategy to the environment it faces. This dynamic responsiveness explains its consistent outperformance across diverse scenarios, as seen both in cost outcomes and pricing behavior trends.

% %%%%%%%%%%%%%%%%%
% %%%%%%%%%%%%%%%%%

\begin{table}[!ht]
\centering
\caption{Average cost of \CSND~under different order-to-customer ratios and its relative savings over benchmark policies.}
\label{table:CSorder_customer_ratio_comparison}
\small
\begin{tabular}{lcccc}
\toprule
\textbf{Order-Customer Ratio} & \textbf{(N+D)} & \textbf{\% Over (N+F)} & \textbf{\% Over (G+D)} & \textbf{\% Over (G+F)} \\
\midrule
\textasciitilde0.33 & \$1,011.31 $\pm$ 2.20 & +2.81\% & +5.99\% & +6.57\% \\
\textasciitilde0.45 & \$1,256.78 $\pm$ 3.10 & +3.01\% & +3.05\% & +4.62\% \\
\textasciitilde0.60 & \$1,470.96 $\pm$ 2.32 & +2.85\% & +3.67\% & +4.81\% \\
\bottomrule
\end{tabular}
\end{table}

% %%%%%%%%%%%%%%%%%

\subsubsection{Variation of Delay Allowed Time}

We now examine how performance varies across different levels of delay flexibility, ranging from 30 to 120 minutes. Table~\ref{table:CSdelay_flexibility_comparison} summarizes the total cost incurred under each policy as the allowable delay increases. Across all scenarios, \CSND~consistently achieves the lowest overall cost and outperforms all three baselines, with savings increasing from 1.57\% to 4.59\% over \CSNF~and reaching up to 5.45\% compared to \CSGF. This underscores \CSND's ability to effectively leverage additional flexibility to improve decision-making and reduce operational costs. The core of \CSND's advantage lies in how it utilizes the delay window, not just to decide when to match, but to choose the right match. As delay allowances grow, \CSND~learns to defer suboptimal matches when necessary, while also identifying high-quality opportunities earlier. This strategic behavior is reflected in the rising number of delay actions remaining at the time of service, from 1.7 in the 30-minute case to 8.5 in the 120-minute case, indicating that \CSND~is often able to match orders shortly after they arrive. Rather than relying on the full delay horizon, it capitalizes on early, high-value matching opportunities, striking a balance between urgency and efficiency.

In terms of matching cost, \CSND~maintains a stable per-epoch average (from 5.20 to 5.41), while greedy baselines such as \CSGF~see significant increases (e.g., from 5.79 to 6.54). Detour distances tell a similar story: \CSND~achieves substantial reductions (from 4.48 to 3.24), whereas \CSGD~and \CSGF~experience growing inefficiencies. This further highlights \CSND's superior route coordination as delay increases. Although greedy policies sometimes match more orders, e.g., at 120 minutes, \CSGD~matches 3.05 orders per epoch compared to \CSND's 2.91, they do so with greater cost and detour penalties. Moreover, \CSND~accumulates more orders over time than \CSGF~(26.7 vs. 19.6 at 120 minutes), demonstrating a strategic, not reactive, approach to batching and delay. Finally, \CSND~achieves a steady reduction in order loss as delay increases, from 0.42 down to 0.30, while maintaining lower lost-order cost than both \CSNF~and \CSGD. Although \CSGF~achieves the lowest order loss, it does so at the expense of significantly higher matching costs. In contrast, \CSND~achieves a well-balanced outcome: early, efficient matching, minimal detours, and consistent cost savings, solidifying its status as the most effective policy under increasing delay flexibility.

\begin{table}[!ht]
\centering
\caption{Average cost of \CSND~across delay times and relative savings over benchmark policies.}
\label{table:CSdelay_flexibility_comparison}
\small
\begin{tabular}{lcccc}
\toprule
\textbf{Delay Allowed} & \textbf{(N+D)} & \textbf{\% Over (N+F)} & \textbf{\% Over (G+D)} & \textbf{\% Over (G+F)} \\
\midrule
30 Minutes  & \$1,335.63 $\pm$ 3.11 & +1.57\% & +2.83\% & +2.96\% \\
60 Minutes  & \$1,275.91 $\pm$ 3.34 & +2.89\% & +3.81\% & +5.21\% \\
90 Minutes  & \$1,256.78 $\pm$ 3.10 & +3.01\% & +3.05\% & +4.62\% \\
120 Minutes & \$1,231.33 $\pm$ 1.90 & +4.59\% & +4.48\% & +5.45\% \\
\bottomrule
\end{tabular}
\end{table}

\subsection{Algorithmic Performance}
\label{sec:CSalgperformance}

Table~\ref{table:CStiming_blocks_comparison} presents a breakdown of computation time per decision epoch across major system components, comparing two settings with differing numbers of unique locations to visit. The most time-consuming operations are concentrated in three key blocks: obtaining values for each feasible action (Block 6), setting up the integer linear program (ILP) (Block 7), and solving the ILP (Block 8). Together, these three stages account for more than 50\% of total runtime in both location settings. As the number of unique locations able to be visited increases, we observe a notable shift in where computational effort is spent. In the setting with fewer locations, computing action values (Block 6) accounts for 18.97\% of the total time. This rises significantly to 40.46\% when more locations are present, reflecting the combinatorial growth of feasible matches that must be evaluated. Similarly, time spent identifying feasible actions (Block 4) and collecting experience (Block 5) grows sharply, indicating that with more destinations, the matching and evaluation space expands, and more complex bookkeeping is required. ILP setup and solve times (Blocks 7 and 8) also increase moderately, from 26.72\% to 29.40\% and 6.80\% to 11.85\%, respectively, in line with the larger decision space introduced by more destinations. Overall, the results indicate that runtime is dominated by neural network inference (Blocks 3 and 6) and optimization (Blocks 7 and 8), with the balance of computation shifting as the number of unique locations to visit grows.

\setlength{\tabcolsep}{6pt}
\renewcommand{\arraystretch}{1.05}
\begin{table}[!ht]
\centering
\caption{Breakdown of time spent per decision epoch across system blocks for two capacity settings.}
\label{table:CStiming_blocks_comparison}
\resizebox{0.85\textwidth}{!}{
\begin{tabular}{c l c c}
\toprule
\textbf{Block} & \textbf{Description} & \textbf{1 Unique Location (\% of Total)} & \textbf{2 Unique Locations (\% of Total)} \\
\midrule
1  & Load new orders and customers                      & 0.003\%  & 0.002\% \\
2  & Initialise new data (set deadlines, etc.)          & 0.054\%  & 0.058\% \\
3  & Get order multipliers for each order               & 8.789\%  & 3.389\% \\
4  & Get feasible actions                               & 5.770\%  & 21.602\% \\
5  & Collect and store experience                       & 7.452\%  & 17.285\% \\
6  & Get values for each feasible action                & 18.970\% & 40.455\% \\
7  & Set up ILP                                         & 26.719\% & 29.395\% \\
8  & Solve ILP                                          & 6.802\%  & 11.847\% \\
9  & Execute decisions                                  & 2.105\%  & 1.093\% \\
10 & Store statistics                                   & 0.100\%  & 0.088\% \\
\bottomrule
\end{tabular}
}
\end{table}

%%%%%%%%%%%%%%%%%%%%%%%%%%%%%%%%%%%%%%%%%
%%%%%%%%%%%%%%%%%%%%%%%%%%%%%%%%%%%%%%%%%
\subsection{Discussions and Managerial Insights} \label{discussion}
The following insights illustrate how the gains from our models can be realized in practice:
\begin{itemize}
    \item Joint optimization through NeurADP + DDQN yields 6.7\% cost savings over NeurADP + Fixed and approximately 2\% over greedy baselines when deviation-time payments are absent. However, when deviation costs rise to \$0.30 per minute, the advantage over NeurADP + Fixed narrows to 0.9\%, though savings over greedy rules increase to approximately 18\%. This suggests that dynamic pricing adds the most value when the system retains flexibility in compensation. Under high-penalty conditions, intelligent matching alone captures most available savings.

    \item Extending the allowable delivery delay from 30 to 120 minutes reduces total cost for the NeurADP + DDQN policy by about 8\% (from \$1,335 to \$1,231). It also increases the margin over NeurADP + Fixed from 1.6\% to 4.6\% and improves performance relative to the best greedy policy. The policy selectively defers only those orders that benefit from waiting, while executing efficient pairings early. This shows that modest slack in delivery deadlines can lower costs without sacrificing performance, as long as scheduling remains forward-looking.

    \item Allowing each crowd-shipper to serve two destinations instead of one reduces daily cost by 17\% (to \$1,036) for the NeurADP + DDQN policy, making it nearly 10\% cheaper than the matching-only variant and about 15\% cheaper than either greedy baseline. Average bundle size nearly doubles, and detour time is cut by more than half. Heuristic matchers, by contrast, show minimal gains. These results underscore that routing flexibility delivers substantial value, particularly when paired with a predictive, optimization-based assignment strategy.

    \item As the order-to-shipper ratio increases from 0.33 to 0.60, total cost rises across all policies. Still, the NeurADP + DDQN policy maintains a consistent advantage of about 3\% over NeurADP + Fixed and up to 6\% over greedy baselines. It achieves this by dynamically adjusting deviation multipliers from 0.8 toward the 1.0 to 1.2 range in response to tightening supply, helping sustain acceptance rates. Fixed-price systems lack this adaptability under changing conditions, making price flexibility essential for maintaining service reliability and cost control.

    \item When the base pay for crowd-shippers is set at \$2, the NeurADP + DDQN policy achieves over 10\% cost savings compared to NeurADP + Fixed and 13 to 15\% over greedy approaches. However, increasing base pay to \$5 reduces these gains, with savings over NeurADP + Fixed falling to around 5\%. These results indicate that to unlock the full potential of dynamic pricing and matching, a meaningful share of compensation must remain discretionary, enabling adaptive incentive design through deviation pay or performance bonuses.
\end{itemize}

%%%%%%%%%%%%%%%%%%%%%%%%%%%%%%%%%%%%%%%%%
%%%% Subsection: Conclusion %%%%
%%%%%%%%%%%%%%%%%%%%%%%%%%%%%%%%%%%%%%%%%
\section{Conclusion} \label{CSconclusion}

% This paper advances the centralized crowd-shipping literature by jointly optimizing order-to-shopper matching and dynamic pricing through a novel NeurADP and DDQN framework. By relaxing limiting assumptions from prior work, such as fixed compensation structures, deterministic acceptance behavior, and single-destination deliveries, we model a more realistic and stochastic urban logistics environment. Our results demonstrate that dynamic pricing leads to meaningful cost reductions, especially when compensation and routing are made more flexible. The proposed policy consistently outperforms fixed pricing approaches and baseline heuristics across a variety of operational settings. Allowing for multiple delivery destinations and more lenient delivery delay windows further enhances efficiency by reducing overall costs and improving route flexibility. These findings highlight the potential of forward-looking, adaptive strategies in last-mile logistics, demonstrating that even moderate operational flexibility, when guided by intelligent decision-making, can drive substantial efficiency gains. Future work may explore expanding pricing strategies by adopting continuous pricing spaces. Currently, deviation pay is limited to a discrete set of multipliers, which simplifies learning but may miss more optimal incentives. A natural next step is to frame pricing as a continuous action problem, using actor–critic or policy-gradient methods to enable more precise and effective incentive design.

This paper contributes to the crowd-shipping with in-store customers literature by jointly optimizing order-to-shopper matching and dynamic pricing through a novel NeurADP and DDQN framework. By relaxing limiting assumptions from prior work, such as fixed compensation structures, deterministic acceptance behavior, and single-destination deliveries, we model a more realistic and stochastic urban logistics environment. Our experiments reveal that incorporating dynamic pricing and operational flexibility leads to substantial cost savings under a range of conditions. 

Future work may explore extending pricing strategies by adopting continuous action spaces. Currently, deviation pay is drawn from a discrete set of multipliers, which simplifies learning but may overlook more precise incentives. Framing pricing as a continuous action problem, using actor-critic or policy-gradient methods, represents a natural next step for enabling fine-tuned, data-driven incentive mechanisms. Another natural extension is to explicitly model the fallback delivery mechanism, which is currently abstracted as a flat-penalty courier. By introducing a second resource class representing this fallback fleet, the system could make real-time trade-offs between in-store crowd-shippers and paid couriers. This extension further opens the door to modeling a heterogeneous paid delivery fleet, such as standard delivery vans or lightweight drones, operating in tandem with the in-store crowd-shippers to fulfill orders.

\singlespacing
\bibliographystyle{elsarticle-harv}
\bibliography{adp_paper}

\end{document}